\documentclass{osa-article}
\usepackage{subfigure}

\usepackage{color,colortbl,soul}
\definecolor{Gray}{gray}{0.9}           
\journal{boe}

\begin{document}

\title{Graph- and finite element-based total variation models for the inverse problem in diffuse optical tomography}

\author{WENQI LU,\authormark{1} JINMING DUAN,\authormark{1} DAVID ORIVE-MIGUEL,\authormark{2,3} LIONEL HERVE,\authormark{2} and IAIN B. STYLES,\authormark{1,*}}

\address{\authormark{1}School of Computer Science, University of Birmingham, UK\\
\authormark{2}CEA, LETI, MINATEC Campus, F-38054 Grenoble, France\\
\authormark{3}Univ. Grenoble Alpes, CNRS, Grenoble INP, GIPSA-lab, 38000 Grenoble, France}

\email{\authormark{*}I.B.Styles@bham.ac.uk} 



\begin{abstract}
Total variation (TV) is a powerful regularization method that has been widely applied in different imaging applications, but is difficult to apply to diffuse optical tomography (DOT) image reconstruction (inverse problem) due to unstructured discretization of complex geometries, non-linearity of the data fitting and regularization terms, and non-differentiability of the regularization term. We develop several approaches to overcome these difficulties by: i) defining discrete differential operators for TV regularization using both finite element and graph representations; ii) developing an optimization algorithm based on the alternating direction method of multipliers (ADMM) for the non-differentiable and non-linear minimization problem; iii) investigating isotropic and anisotropic variants of TV regularization, and comparing their finite element- and graph-based implementations. These approaches are evaluated on experiments on simulated data and real data acquired from a tissue phantom. Our results show that both FEM and graph-based TV regularization is able to accurately reconstruct both sparse and non-sparse distributions without the over-smoothing effect of Tikhonov regularization and the over-sparsifying effect of L$_1$ regularization. The graph representation was found to out-perform the FEM method for low-resolution meshes, and the FEM method was found to be more accurate for high-resolution meshes.
\end{abstract}

\section{Introduction}
Diffuse optical tomography (DOT) is an important non-invasive imaging technique whose major applications include diagnosing breast cancer \cite{gibson2005recent,srinivasan2003interpreting,dehghani2003multiwavelength}, imaging small animals for the study of disease and analyzing brain function in functional neuroimaging \cite{boas2004improving,custo2010anatomical,niu2011resting,eggebrecht2014mapping}. In DOT, near-infrared light (spectral range of 650-900 nm) is injected into the object through optical fibers placed on the surface of the object. The transmitted light is then collected using optical detectors, forming a series of boundary measurements, each of which corresponds to the signal received by a single detector during illumination by a single source. Image reconstruction algorithms are then used to recover the internal distribution of the underlying optical properties of the object from the boundary measurements. 

Due to the limited availability of boundary measurements and diffusive nature of near-infrared light propagation, image reconstruction in DOT is an underdetermined, ill-posed and non-linear inverse problem. Regularization is often used to constrain the inverse problem to yield physiologically and anatomically plausible solutions, resulting in the following minimization problem with respect to the optical properties $\mu$

\begin{equation} \label{eq:inverse11}
{\mu ^*} = \arg {\min _\mu }\left\{ \frac{1}{2}\| {\Phi _{}^{\rm{M}} - {\cal F}\left( \mu  \right)\|_2^2 + \lambda {\cal R}\left( \mu  \right)} \right\},
\end{equation}
where $\Phi^{\rm{M}}$ represents the boundary measurements acquired from the optical detectors, ${\cal F}$ is the non-linear operator induced from the forward model \cite{arridge1997optical}, ${\cal R}$ is a general regularization term, and $\lambda$ is a weight that determines the extent to which regularization will be imposed on the solution $\mu^*$. The quadratic Tikhonov-type regularization is widely used. However, it promotes smooth solutions and thereby smears sharp features embedded in the image \cite{yalavarthy2007structural}. Regularizations based on the L$_1$-norm of the solution have been also extensively studied \cite{shaw2012effective,baritaux2011sparsity,kavuri2012sparsity}, as they impose a sparsity constraint on the solution, enabling the recovery of sharp edges of objects in reconstructed images. Eq. (\ref{eq:inverse11}) with either regularization mentioned results in a convex optimization problem, where highly efficient algorithms \cite{lu20181} are available. Recently, the more general L$_p$ regularization ($0 < p < 1$) that promotes sparsity into the resulting image \cite{lyu2013comparison}, has been employed for DOT image reconstruction \cite{prakash2014sparse,okawa2011improvement}. However, L$_p$ regularization is nonconvex and therefore difficult to optimize. 

Regularizations involving the L$_1$ or L$_p$-norm of the solution are used under the assumption that the optical properties (representing the image) to be reconstructed are spatially sparse. These regularizations tend to oversparsify the distribution of the optical properties when such an assumption does not hold \cite{lu20181}, for example, in the case of multiple activations or complex injuries in the brain, where the features of interest are not spatially localized and the optical properties relative to the background are therefore non-sparse \cite{eggebrecht2014mapping}. In order to be able to reconstruct images in which edges are preserved and features are not spatially sparse, a different approach is required. Total variation (TV) regularization, which uses the L$_1$-norm of the \textit{gradient} of the solution as a regularizing term (the detailed forms will be given in Section \ref{eq:FETV} and  \ref{eq:GTV}), can be used to overcome the limitations. The gradient operator can transform the solution $\mu^*$ to a sparse space where non-zero values only occur at sharp features. As such, TV can perform better than the pure sparsity preserving regularizations at preserving edges of objects in the images that are not sparse. Further, gradient is a highpass operator, which imposes smoothness to the solution. This improves the conditioning of the minimization problem (Eq. \eqref{eq:inverse11}), thus enabling a robust numerical solution. Due to these advantages, TV has been adapted from applications in imaging processing  \cite{osher2005iterative,duan2015fast,duan2016denoising} to various medical image reconstruction problems, including photoacoustic tomography (PAT) \cite{yao2011enhancing}, bioluminescence tomography (BLT) \cite{gao2010multilevel}, fluorescence tomography (FT) \cite{freiberger2010total}, as well as DOT \cite{konovalov2016total,paulsen1996enhanced,tang2017mixed}.

In most TV-associated imaging problems, the minimization problem is carried out on a Cartesian grid where each element represents a pixel (voxel in 3D) in the image \cite{lu2016implementation}. In this case, the differential operators resulting from minimizing the TV regularization, such as gradient, divergence, Laplacian and curvature, are discretized straightforwardly using the finite difference method (FDM) \cite{lu2016implementation}. In DOT, it is however non-trivial to represent the complex geometry (i.e. the multi-layer head used in our experiment) using a Cartesian grid and the FDM is not always practical. Two representations are often employed to model complex geometries: finite element and graph representations. In the former, the object geometry is represented by a polygon/polyhedron, over which a series of disjoint triangles (tetrahedron in 3D) are usually generated. In the latter, the object geometry is represented by an unstructured graph, defined by vertices, edges and weights. Such a graph is normally constructed by exploring neighborhood relationships between vertices. For each representation, there is a systematic discretization scheme (finite element discretization or graph discretization) for the differential operators, which can be readily applied to the minimization of TV-associated problems. We note that although TV regularization has previously been studied in DOT \cite{konovalov2016total,paulsen1996enhanced,tang2017mixed}, none of these studies have provided detailed information about the discretization schemes they used. The performance of different discretization schemes for TV regularization in DOT has not been systematically investigated.

The minimization of a TV-associated problem can be non-trivial due to the non-linearity and non-differentiability of the TV regularization term. In image processing, many efficient optimization algorithms have been developed for this task, including iteratively reweighed least squares \cite{daubechies2010iteratively}, primal dual \cite{pock2009algorithm}, split Bregman \cite{goldstein2009split}, and fast iterative shrinkage-thresholding algorithm (FISTA) \cite{beck2009fast,goldstein2014field}. Recently, alternating direction method of multipliers (ADMM) \cite{papafitsoros2014combined,duan2015fast,lu2016implementation,duan2016edge} has become increasingly popular. The elegance of ADMM lies in decomposition of the original minimization problem into several simple subproblems, each of which either has a closed-form solution or can be iteratively solved with efficient numerical methods. However, since ADMM-based methods have been implemented mainly for Cartesian grids using a forward-backward FDM \cite{duan2017introducing} and it is not straightforward to generalize them to solve the inverse problem on an unstructured domain. Moreover, the non-linearity of the data fitting term in Eq. \eqref{eq:inverse11} further complicates the DOT reconstruction problem, making the minimization process required to solve Eq. \eqref{eq:inverse11} difficult. 

In this paper, we address these limitations and develop TV regularization approaches for the inverse problem in DOT. More specifically, we make the following three distinct contributions: \textbf{(1)} We introduce finite element and graph representations to discretize the TV regularization term in DOT reconstruction enabling the minimization of the inverse problem associated with TV regularization to be carried out on unstructured domains. To the best of our knowledge, this is the first time that finite element-based discretization methods have been provided in detail for DOT image reconstruction with TV regularization. Additionally, we are not aware of any previous work that attempts to formulate the TV-regularized inverse problem using a graph representation. \textbf{(2)} We propose an efficient algorithm based on ADMM to minimize the TV-regularized inverse problem. Our algorithm can handle geometries with unstructured grids, and also reduces the computational difficulties arising from the non-differentiability and dual non-linearities in the inverse problem. \textbf{(3)} We further investigate the isotropic and anisotropic variants of the TV regularization, and compare their finite element- and graph-based implementations against a baseline Tikhonov model, both qualitatively and quantitatively using extensive numerical experiments.

\begin{figure}[ht]
\centering
\includegraphics[width=0.7\textwidth]{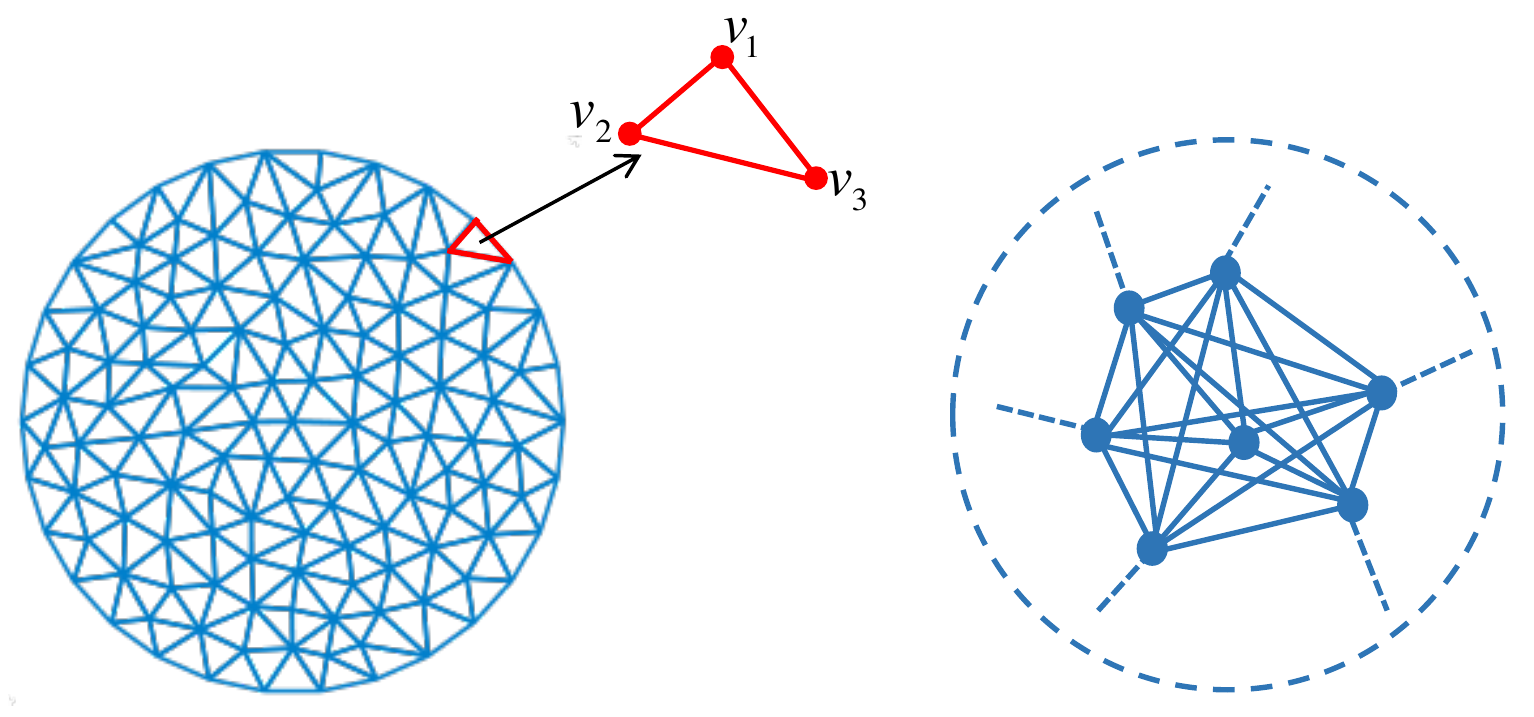}
\caption{Modeling a complex geometry using finite element (left) and graph (right) representations. }
\label{fig:p02}
\end{figure}

\section{Discretizations using finite element and graph representations}
In this section, we first show how an unstructured computational domain can be modelled using finite element (FE) and graph representations. The definitions of TV regularizations (anisotropic and isotropic versions) and corresponding discrete differential operators are then given accordingly for each representation. In Fig. \ref{fig:p02}, we show a circle discretized with the two representations. For the FE representation (left), the circle is divided by a series of elements joint at different vertices. In the FE representation, we normally term a discrete geometry as a FE mesh. Within the circle mesh, a triangle (representing one element) is highlighted comprising three disjoint vertices. For the graph representation (right), the circle is simply discretized with a set of vertices and edges. In this representation there is no concept of `element'. Note that it is easy to convert the FE representation to the graph representation. For example, the FE mesh can be viewed as a graph if we consider only the vertices and edges in it. In this section, two representations are introduced to model the unstructured computational domain of complex DOT geometries.

\subsection{Finite element-based discrete differential operators}
\label{eq:FETV}
We apply the Galerkin FE method to the computational domain in DOT \cite{reed1973triangular}, the first step of which is to approximate a continuous function by a piecewise-polynomial function. Using the FE representation, let us first discretize an unstructured 2D domain $\Omega$ by $M$ triangles jointed at $N$ vertex nodes (e.g. Fig. \ref{fig:p02} left). $V = \left\{ {{V_k}} \right\}_{k = 1}^N$ denotes a finite number of $N$ nodes. Let $\vartheta_h$ be the 2D vector space of continuous piecewise-linear functions on the triangles in the FE mesh. The continuous and piecewise-linear function $U(x, y): \Omega  \to \mathbb{R}$, approximating the optical properties on $\Omega$, can be written in the form of
\begin{equation} \label{eq:FETV1}
{U} = \sum\limits_{i = 1}^N {{\mu_i}} {\varphi _i}.
\end{equation}
Here $\left\{ {{\varphi _i}} \right\}_{i = 1}^{N}$, ${\varphi _i} \in \vartheta_h$ are linear basis functions defined as $\varphi_j(V_i) = 1$ if $i=j$ and $\varphi_j(V_i) = 0$ if $i\neq j$. $\mu_i: V \to \mathbb{R}$ is the value of optical property on each vertex in the FE mesh, $i=1,...,N$. 

Eq. \eqref{eq:FETV1} means that the optical property value inside a triangle is associated with the optical property values on all vertices in the mesh. Given three vertices of a triangle $T$, i.e. $v_{1}= \left ( x_{1},y_{1} \right )$, $v_{2}= \left ( x_{2},y_{2} \right )$ and $v_{3}= \left ( x_{3},y_{3} \right )$, there are three linear basis functions $\varphi _i$ associated with the vertices, which are respectively expressed as
\begin{equation} \label{eq:FETV3}
\begin{gathered}
  {\varphi _1}\left( {x,y} \right) = {a_1}x + {b_1}y + {c_1} \hfill \\
  {\varphi _2}\left( {x,y} \right) = {a_2}x + {b_2}y + {c_2} \hfill \\
  {\varphi _3}\left( {x,y} \right) = {a_3}x + {b_3}y + {c_3} \hfill  \\
\end{gathered} 
: \; \; \Omega \to \mathbb{R},
\end{equation}
$a_1=\left(y_{2}-y_{3}\right)/{(2A_{T})}$, $b_1=\left(x_{3}-x_{2}\right)/{(2A_{T})}$, $c_1=\left(x_{2}y_{3}-x_{3}y_{2}\right)/{(2A_{T})}$, $a_2=\left(y_{3}-y_{1}\right)/{(2A_{T})}$, $b_2=\left(x_{1}-x_{3}\right)/{(2A_{T})}$, $c_2=\left(x_{3}y_{1}-x_{1}y_{3}\right)/{(2A_{T})}$, $a_3=\left(y_{1}-y_{2}\right)/{(2A_{T})}$, $b_3=\left(x_{2}-x_{1}\right)/{(2A_{T})}$ and $c_3=\left(x_{1}y_{2}-x_{2}y_{1}\right)/{(2A_{T})}$. $\left ( x,y \right )$ represents any point inside of the triangle $T$. $A_{T}$ denotes the triangular area of $T$, which is computed as
$A_{T}=\left | x_{1}\left ( y_{2}-y_{3} \right ) + x_{2}\left ( y_{3}-y_{1} \right ) + x_{3}\left ( y_{1}-y_{2} \right )\right |/{2}$.

In FE, one starts from a continuous problem and approximates the solution with a piecewise-polynomial function $U$. As such, we define the following anisotropic and isotropic TV regularizations 

\begin{equation}\label{eq:FETV4}
\int_\Omega  {\left( {\left| {{\partial _x}U} \right| + \left| {{\partial _y}U} \right|} \right)} dxdy = {\left\| {{D_x}\mu } \right\|_1} + {\left\| {{D_y}\mu } \right\|_1} .
\end{equation}
\begin{equation}\label{eq:FETV5}
\int_\Omega  {\sqrt {{{\left( {{\partial _x}U} \right)}^2} + {{\left( {{\partial _y}U} \right)}^2}} } dxdy = \sum\limits_{i = 1}^M {\sqrt {{{\left| {{{\left( {{D_x}\mu } \right)}_i}} \right|}^2} + {{\left| {{{\left( {{D_y}\mu } \right)}_i}} \right|}^2}} } .
\end{equation}
In Eq. \eqref{eq:FETV4} and Eq. \eqref{eq:FETV5}, the continuous TV regularizations and their resulting discretized versions are shown on the left-hand side and right-hand side, respectively. The two discrete versions respectively are the anisotropic and isotropic definitions of TV regularization. $\partial_x$ and $\partial_y$ are continuous partial derivatives along the $x$ and $y$ directions, respectively. $D_x$ is a matrix of size $M \times N$ which, when applied to $\mu$ gives the discrete partial derivative of $\mu$ along the $x$ direction. $D_y$ is the derivative matrix along the $y$ direction. $D_x \mu$ and $D_y \mu$ are therefore two vectors of size $M\times 1$, where $M$ is the number of triangles in the mesh. We note that the main idea of FE is to break down the calculation domain $\Omega$ onto the local elements individually. Afterwards, the derived local matrices are assembled element by element to enable the final computation. Eq. \eqref{eq:FETV4} and Eq. \eqref{eq:FETV5} can be proved by expressing the partial derivatives $\partial_x U$ and $\partial_y U$ in terms of a basis. To illustate this idea we prove the first term of Eq. \eqref{eq:FETV4}:

\begin{equation} \label{eq:FETV6}
\begin{split}
\int_\Omega  {\left| {{\partial _x}U} \right|} dxdy &= \sum\limits_{i = 1}^M {\int_{{T_i}} {\left| {{\partial _x}U} \right|dxdy} }  = \sum\limits_{i = 1}^M {\int_{{T_i}} {| {\sum\limits_{j = 1}^N {{\mu _j}{\partial _x}{\varphi _j}} } |dxdy} } \\ & = \sum\limits_{i = 1}^M {{A_{{T_i}}}\left| {{a_{i,1}}{\mu _{i,1}} + {a_{i,2}}{\mu _{i,2}} + {a_{i,3}}{\mu _{i,3}}} \right|}  \\ & = \sum\limits_{i = 1}^M {\left| {{{\left( {{D_x}\mu } \right)}_i}} \right|}  = {\left\| {{D_x}\mu } \right\|_1},
\end{split}
\end{equation}
where ${A_{{T_i}}}$  denotes the area of triangle $T_i$ and $\left \{ i,1 \right \}$, $\left \{ i,2 \right \}$, $\left \{ i,3 \right \}$ in ${\left| {{a_{i,1}}{\mu _{i,1}} + {a_{i,2}}{\mu _{i,2}} + {a_{i,3}}{\mu _{i,3}}} \right|}$ represent the indices of the vertices of the $ith$ triangle. As ${\left| {{a_{i,1}}{\mu _{i,1}} + {a_{i,2}}{\mu _{i,2}} + {a_{i,3}}{\mu _{i,3}}} \right|}$ is a linear combination, we can thus construct the discrete derivative matrix $D_x$ with the following steps:

\begin{itemize}
\item Initialize all-zeros matrix $D_x$ of size $M \times N$.
\item Loop over $M$ triangles; for each triangle $i$, compute the coefficients $a_1$, $a_2$ and $a_3$ using the coordinates of the three vertices and fill in the three columns in the $ith$ row of matrix $D_x$ corresponding to the position of the three vertices in the node sequence.
\end{itemize}

The discrete derivative matrix $D_y$ can be obtained in a similar way. Note that $D_x$ and $D_y$ are sparse matrices as their most entries are zeros. With $D_x$ and $D_y$ defined, we can therefore minimize the TV regularization (either anisotropic (Eq. \eqref{eq:FETV4}) or isotropic (Eq. \eqref{eq:FETV5}) version) with the data fidelity term in (Eq. \eqref{eq:inverse11}) for DOT reconstruction over 2D unstructured geometries. The corresponding 3D counterparts were also implemented in this paper, as shown in the experiments.   

\subsection{Graph-based discrete differential operators}
\label{eq:GTV}
In this section, we introduce discrete differential operators on graphs and from these derive the TV regularization terms. First, we discretize an unstructured domain $\Omega$ by a weighted graph $G = \left( {V,E,w} \right)$ (e.g. Fig. \ref{fig:p02} right). In the graph $G$, $V = \left\{ {{V_k}} \right\}_{k = 1}^N$ denotes a finite set of $N$ vertices, and $E \in V \times V$ is a finite set of weighted edges. We assume that $G$ is an undirected simple graph (no multiple edges) in this study. Let $(i,j) \in E$ be an edge of $E$ that connects the vertices $i$ and $j$ in $V$. Let $\mu_i: V \to \mathbb{R}$ denote the value of the optical properties on ${i}$. The DOT reconstruction problem then reduces to finding an optimal value of the optical properties for each of $N$ vertices in $V$.

The discrete differential operators are defined on the graph  based on nonlocal methods \cite{gilboa2008nonlocal,duan2015fast,duan2013color}.  First, we define the \textit{nonlocal gradient operator} $\nabla_w$ acting on $\mu_i$
\begin{equation} \label{eq:nonlocalgradient}
{\nabla_w}{\mu _{i}} \triangleq \left( {{\mu _j} - {\mu _i}} \right)\sqrt {{w_{ij}}} {\text{       }} :V \to \mathbb{R}.
\end{equation}

For vertex $i \in V$, ${\nabla_w}{\mu _{i}}$  is a vector with a length of $N$. The weight $w_{i,j}: V \times V \ \to \mathbb{R}^+$ represents the similarity between nodes $i$ and $j$, which is nonnegative and symmetric. The weight function can be determined in many ways. In this study we choose to use the Euclidean distance to define the weight function with $w_{i,j}=1/d_{i,j}$ where $d_{i,j}$ represents the distance between vertex $i$ and $j$. We note that an important difference between the FE gradient and the nonlocal gradient is that the former has two directions in 2D or three directions in 3D, whilst the latter is a vector of  partial derivatives along all edges connected to the node.

Given a vector function $\boldsymbol{\nu} _{i} : V \to \mathbb{R}$, the \textit{nonlocal divergence operator} $\mathrm{div}_w$ acting on $\boldsymbol {\nu} _{i}$ is given as
\begin{equation} \label{eq:nonlocaldivergence}
\mathrm{div}_w\, \boldsymbol{\nu} _i \triangleq \sum\limits_{j=1}^N {\left( {{\nu _{ij}} - {\nu _{ji}}} \right)\sqrt {{w_{ij}}} } {\text{ }}: V \to \mathbb{R}
,\end{equation}
where ${\nu _{ij}}$ is the $jth$ index in the vector $\boldsymbol{\nu} _{i}$. 

Based on Eq. \eqref{eq:nonlocalgradient} and Eq. \eqref{eq:nonlocaldivergence}, the \textit{nonlocal Laplace operator} $\Delta_w$ acting on $\mu_i$ is written as
\begin{equation} \label{eq:nonlocallap}
{\Delta_w}{\mu _i} \triangleq \frac{1}{2}\mathrm{div}_w\,\left( {{\nabla}_w{\mu _i}} \right) = \sum\limits_{j =1}^N {\left( {{\mu _j} - {\mu _i}} \right){w_{ij}}}: V \to \mathbb{R},
\end{equation}
which is a linear operator also known as the \textit{graph Laplacian}.

With these discrete differential operators defined on graph, we propose the anisotropic graph TV
\begin{equation} \label{eq:GTV-1}
\left\| {\nabla_w} \mu  \right\|_1 = \sum\limits_{i =1}^{N} {\sum\limits_{j =1}^{N} {\left| {\left( { {\mu _j} -  {\mu _i}} \right)\sqrt {{w_{ij}}} } \right|} },
\end{equation}
and the isotropic graph TV regularization
\begin{equation} \label{eq:GTV-2}
\left\| {\nabla_w} \mu  \right\|_1 = \sum\limits_{i=1}^{N} {\sqrt {\sum\limits_{j =1}^{N} {{{\left( { {\mu _j} -  {\mu _i}} \right)}^2}{w_{ij}}} } }.
\end{equation}
In the above definitions, the constructed graph $G$ is assumed to be fully connected, meaning that each vertex is connected to all other vertices in $G$. In this case, the computational load for the minimizations of Eq. \eqref{eq:GTV-1} and Eq. \eqref{eq:GTV-2} will be extremely heavy. Spectral graph theory \cite{bertozzi2012diffuse,merkurjev2013mbo} or nearest neighbour \cite{elmoataz2008nonlocal,bresson2014multi} techniques are typically employed to limit the number of edges that are considered. For example, \cite{bertozzi2012diffuse} and \cite{merkurjev2013mbo} use spectral approaches and the Nystrom extension method \cite{fowlkes2004spectral} to efficiently calculate the eigen-decomposition of a dense graph Laplacian. In this work, we build the graph by borrowing the positions of the vertices and the connectivity between vertices in the finite element mesh as the vertices and edges in the graph. With this structure, the graph is sparsified and only connected vertices for a given vertice are taken into consideration. In such a case, Eq. \eqref{eq:GTV-1} and Eq. \eqref{eq:GTV-2} are equivalent to 
\begin{equation} \label{eq:GTV-A}
\left\| {\nabla_w} \mu  \right\|_1 = \sum\limits_{i =1}^{N} {\sum\limits_{j \in {\cal N}_i} {\left| {\left( { {\mu _j} -  {\mu _i}} \right)\sqrt {{w_{ij}}} } \right|} },
\end{equation}
and 
\begin{equation} \label{eq:GTV-I}
\left\| {\nabla_w} \mu  \right\|_1 = \sum\limits_{i=1}^{N} {\sqrt {\sum\limits_{j \in {\cal N}_i} {{{\left( { {\mu _j} -  {\mu _i}} \right)}^2}{w_{ij}}} } },
\end{equation}
where ${{\cal N}_i} = \left\{ {j \in V:\left( {i,j} \right) \in E} \right\}$. We note that the 2D and 3D implementations of these differential operators are identical, making the resulting minimization processes of Eq. \eqref{eq:GTV-A} and Eq. \eqref{eq:GTV-I} more straightforward than the FE implementation.

\section{Minimization of TV-associated DOT inverse problems}
\label{minimization}
Due to the non-linearity of the data fitting term and the non-differentiability of the TV regularizations, it is non-trivial to minimize a TV-regularized inverse problem. It is harder than minimizing the standard L1-regularized inverse problem \cite{lu20181} because of the existence of the gradient operator. In this section, we propose an efficient algorithm based on ADMM to address this, the idea of which is to first linearize the non-linear inverse problem and afterwards apply ADMM to the resulting linearized problem. The whole process is then iterated until convergence. We note that due to the use of the differential operators in Sections \ref{eq:FETV} and \ref{eq:GTV}, the proposed algorithm can handle complex geometries with unstructured grids, and also can ease the computational difficulties arising from the non-differentiability and non-linearities in the inverse problem. We now describe the details of this algorithm.

\subsection{Linearization}
Non-linear problems are technically difficult to tackle directly, so iterative linearization can be used to convert a non-linear problem into a series of local linear problems. To do so, Taylor's series expansion is first used to approximate ${\cal F}(\mu)$ in the fitting term of Eq. \eqref{eq:inverse11} as
\begin{equation} \label{eq:Linear}
{\cal F}\left ( \mu  \right )\approx {\cal F}\left ( \mu ^{k-1} \right )+{J}^{k-1}\left ( \mu - \mu^{k-1} \right ),
\end{equation}
where $J$ denotes the Jacobian calculated from the $(k-1)$th iteration and is defined as ${{\partial {\cal F}({\mu ^{k - 1}})} \mathord{\left/{\vphantom {{\partial F({\mu ^{k - 1}})} {\partial ({\mu ^{k - 1}})}}} \right. \kern-\nulldelimiterspace} {\partial ({\mu ^{k - 1}})}}$. The Jacobian in DOT is normally calculated using the adjoint method \cite{arridge1995photon2}. With this first order approximation, the non-linear problem Eq. \eqref{eq:inverse11} can be converted to
\begin{equation} \label{eq:twostepLinears}
{\Delta\mu ^k} = \arg {\min _{\Delta\mu} }\left\{  \frac{1}{2} \|{J^{k - 1}}{\Delta\mu}  - {\Delta\Phi^{k-1}}\|_2^2 + \lambda {\cal R}\left( \Delta\mu  \right) \right\},
\end{equation}
which is an iterative linearized algorithm. $\Delta \Phi^{k-1}$ is the data-model mismatch which is given by ${\Phi ^{\rm{M}}} - {\cal F}\left( {{\mu ^{k - 1}}} \right)$ and $\Delta\mu^k$ is the change in the optical property at the $k$-th iteration. With a proper initialization $\mu^0$, a local minimizer to Eq. \eqref{eq:inverse11} can be found by iteratively updating this step. In Eq. \eqref{eq:FETV4}, Eq. \eqref{eq:FETV5}, Eq. \eqref{eq:GTV-A} and Eq. \eqref{eq:GTV-I}, we have defined the four types of TV regularizations using different representations. We then apply them to ${\cal R}(\Delta\mu)$ in Eq. \eqref{eq:twostepLinears}, resulting in the following four linearized minimization problems in Table~\ref{tb:summeriseModels}. However, it remains unclear how to optimize the second step in Eq. \eqref{eq:twostepLinears}. We address this using ADMM. 

\newcolumntype{L}{>{\centering\arraybackslash}m{3cm}}
\begin{table}[ht]
\centering 
\caption{Four TV-regularized minimization problems obtained by applying different TV regularizations to Eq. \eqref{eq:twostepLinears}. A-FETV, I-FETV, A-GTV and I-GTV respectively represent anisotropic finite element total variation,  isotropic finite element total variation, anisotropic graph total variation and isotropic graph total variation.}
\label{tb:summeriseModels}
\resizebox{0.8\columnwidth}{!}{
\vspace{-10pt}
\begin{tabular}{|L|c|} 
\hline
 Name & Formulation\\\hline
 A-FETV & \footnotesize{${\Delta\mu ^*} = \arg {\min \limits_{\Delta\mu} }\left\{ \frac{1}{2} \|{J{\Delta\mu}  - {\Delta\Phi}\|_2^2 + \lambda {{\left\| {{D_x}(\Delta\mu) } \right\|}_1} + \lambda {{\left\| {{D_y}(\Delta\mu) } \right\|}_1}} \right\}$} \\
\hline
 I-FETV &\footnotesize${\Delta\mu ^*} = \arg {\min \limits_{\Delta\mu} }\left\{ \frac{1}{2} \|{J}{\Delta\mu}  - {\Delta\Phi}\|_2^2 +\lambda \sum\limits_{i = 1}^M {\sqrt {{{\left| {{{\left( {{D_x}(\Delta\mu) } \right)}_i}} \right|}^2} + {{\left| {{{\left( {{D_y}(\Delta\mu) } \right)}_i}} \right|}^2}}} \right\}
$\\
\hline
 A-GTV & \footnotesize${\Delta\mu ^*} = \arg {\min \limits_{\Delta\mu} }\left\{ \frac{1}{2} \|{{J}{\Delta\mu}  - {\Delta\Phi}\|_2^2 + \lambda \sum\limits_{i =1}^N  {\sum\limits_{j \in {{\cal N}_i}}^{} {\left| {\left( {{{\Delta\mu} _j} - {{\Delta\mu} _i}} \right)\sqrt {{w_{ij}}} } \right|} } } \right\}$\\
\hline
 I-GTV & \footnotesize${\Delta\mu ^*} = \arg {\min \limits_{\Delta\mu}}\left\{ \frac{1}{2} \|{{J}{\Delta\mu}  - {\Delta\Phi}\|_2^2 + \lambda \sum\limits_{i =1}^N {\sqrt {\sum\limits_{j \in {\mathcal{N}_i}}^{} {{{\left( { {{\Delta\mu} _j} -  {{\Delta\mu} _i}} \right)}^2}{w_{ij}}} } } } \right\}$\\
\hline
\end{tabular}
}
\end{table} 

\subsection{ADMM Implementations}
In this section, we introduce ADMM in detail to minimize the A-FETV, I-FETV, A-GTV and I-GTV problems. We begin with the ADMM implementation for A-FETV. Specifically, auxiliary splitting vectors ${\nu}_x$ and ${\nu}_y$ are introduced to represent ${D_x}(\Delta\mu)$ and ${D_y}(\Delta\mu)$ respectively. Therefore the A-FETV problem is transformed into the following unconstrained optimization problem:
\begin{equation} \label{eq:Ani1}
\begin{split}
{{\Delta\mu}^n,{\nu}_x^n,{\nu}_y^n} = \mathop {\arg \min }\limits_{{\Delta\mu},{\nu}_x,{\nu}_y} &  \{ \frac{1}{2}||{J}{\Delta\mu}  - {\Delta\Phi} ||_2^2 + \lambda ||{\nu}_x |{|_1} + \lambda ||{\nu}_y |{|_1} \\ & + \frac{\theta }{2}|| {\nu}_x - {D_x}(\Delta\mu) - {b}_x^{n-1}||_2^2 + \frac{\theta }{2}|| {\nu}_y - {D_y}(\Delta\mu) - {b}_y^{n-1}||^2_2 \} ,
\end{split}
\end{equation}
where superscript $n$ denotes the $n$th ADMM iteration and ${b}_x$ and ${b}_y$ are iterative parameters. In order to find the minimizer of Eq. \eqref{eq:Ani1}, an alternating optimization method is used where Eq. \eqref{eq:Ani1} is split into several subproblems with respect to ${\Delta\mu}$, ${\nu}_x$, ${\nu}_y$, ${b}_x$ and ${b}_y$, each of which can be solved separately.

First the iterative minimization approach requires us to solve the subproblem with respect to $\mu$
\begin{equation} \label{eq:Ani4}
\begin{split}
{{\Delta\mu}^{n}} = \mathop {\arg \min }\limits_{\Delta\mu}  \{ \frac{1}{2}||{J}{\Delta\mu}  - {\Delta\Phi} ||_2^2 & + \frac{\theta }{2}|| {\nu}_x^{n-1} - {D _x}(\Delta\mu) - b_x^{n-1}||_2^2 \\ & + \frac{\theta }{2}|| {\nu}_y^{n-1} - {D_y}(\Delta\mu) - b_y^{n-1}||_2^2\},
\end{split}
\end{equation}
which has the optimality condition
\begin{equation} \label{eq:Ani5}
\left ( \left ( J^{T}J+\theta \left ( {D _x}^{T}{D _x}+{D_y}^{T}{D_y} \right ) \right ) \right ){\Delta\mu}^{n}={J}^{T}{\Delta\Phi} -\theta {D _x}^{T}\left ( b_x^{n-1}-{\nu}_x^{n-1} \right ) -\theta {D_y}^{T}\left ( b_y^{n-1}-{\nu}_y^{n-1} \right ) .
\end{equation}

As the inversion matrix of Eq. \eqref{eq:Ani5} has size  $N \times N$, in order to achieve high efficiency, we use a gradient descent method to optimize the functional iteratively, in which the step size controls how far the iterate moves along the gradient direction during the current iteration. Instead of setting the step size manually, we use a backtracking line search to enforce convergence \cite{goldstein2014field}. 

The next subproblem with respect to ${\nu}_x$ and ${\nu}_y$ is given as
\begin{equation} \label{eq:sb2}
\begin{split}
{{\nu}_x^n,{\nu}_y^n} = \mathop {\arg \min }\limits_{{\nu}_x,{\nu}_y}  \{ \lambda ||{\nu}_x |{|_1} + \lambda ||{\nu}_y |{|_1} & + \frac{\theta }{2}|| {\nu}_x - {D_x}(\Delta\mu^n) - b_x^{n-1}||_2^2 \\ & + \frac{\theta }{2}|| {\nu}_y - {D_y}(\Delta\mu^n) - b_y^{n-1}||_2^2 \}. 
\end{split}
\end{equation}
It should be noticed that, in A-FETV, there is no coupling between ${\nu}_x$ and ${\nu}_y$. We can explicitly compute the optimal value of ${\nu}_x$ and ${\nu}_y$ using the generalized shrinkage operators
\begin{equation} \label{eq:Ani6}
\begin{gathered}
  {\nu}_{x}^{n}={\rm{max}}\left ( \left | D_x (\Delta\mu^{n})+b_x^{n-1} \right |-\frac{\lambda} {\theta},0 \right ) \frac{D_x (\Delta\mu^{n})+b_x^{n-1}}{\left | D_x (\Delta\mu^{n})+b_x^{n-1} \right |}   \hfill \\
  {\nu}_{y}^{n}={\rm{max}}\left ( \left | D_y (\Delta\mu^{n})+b_y^{n-1} \right |-\frac{\lambda} {\theta},0 \right ) \frac{D_y (\Delta\mu^{n})+b_y^{n-1}}{\left | D_y (\Delta\mu^{n})+b_y^{n-1} \right |}, \hfill \\ 
\end{gathered} 
\end{equation}
with the convention that $0/0=0$. The last one is to update the iterative parameters $b_x$ and $b_y$, as
\begin{equation} \label{eq:Ani8}
\begin{gathered}
b_x^{n}=b_x^{n-1}+D_x (\Delta\mu^{n})-{\nu}_x^{n}  \hfill \\
b_y^{n}=b_y^{n-1}+D_y (\Delta\mu^{n})-{\nu}_y^{n} . \hfill \\ 
\end{gathered} 
\end{equation}

In I-FETV, using the same alternating optimization method, the original minimization problem can be transformed as

\begin{equation} \label{eq:Ani10}
\begin{split}
{{\Delta\mu}^n,{\nu}_x^n,{\nu}_y^n} = \mathop {\arg \min }\limits_{{\Delta\mu},{\nu}_x,{\nu}_y}  \{ & \frac{1}{2}||{J}{\Delta\mu}  - {\Delta\Phi} ||_2^2 + \lambda \left \| \left ( {\nu}_x,{\nu}_y \right ) \right \|_2 \\ & + \frac{\theta }{2}|| {\nu}_x - {D _x}(\Delta\mu) - b_x^{n-1}||^2_2 + \frac{\theta }{2}|| {\nu}_y - {D _y}(\Delta\mu) - b_y^{n-1}||_2^2 \} ,
\end{split}
\end{equation}
where
\begin{equation} \label{eq:Ani11}
\left \| \left ( {\nu}_x,{\nu}_y \right ) \right \|_2=\sum_{i=1}^{M} \sqrt{\left| \left ( {\nu}_x \right )_i \right| ^2+\left| \left ( {\nu}_y \right )_i \right|^2},
\end{equation}
and $M$ is the number of finite elements. The first subproblem (L$_2$ component) with respect to $\Delta\mu$ is the same as A-FETV. It should be noted that the ${\nu}_x$ and ${\nu}_y$ variables cannot be decoupled as they were in A-FETV. In order to solve the subproblem with respect to ${\nu}_x$ and ${\nu}_y$, we can explicitly solve the minimization problem for $\left ( {\nu}_x^n,{\nu}_y^n \right )$, using a generalized shrinkage formula
\begin{equation} \label{eq:Ani13}
\begin{gathered}
{\nu}_{x}^{n} = \max \left( {{s^n} - \frac{\lambda}{\theta} ,0} \right)\frac{D_x (\Delta\mu^{n})+b_x^{n-1}}{{{s^n}}}  \hfill \\
{\nu}_{y}^{n} = \max \left( {{s^n} - \frac{\lambda}{\theta} ,0} \right)\frac{D_y (\Delta\mu^{n})+b_y^{n-1}}{{{s^n}}} , \hfill \\ 
\end{gathered}
\end{equation}
with the convention that $0/0=0$ and ${s^n} = \sqrt {{{\left| {D_x (\Delta\mu^{n})+b_x^{n-1}} \right|}^2}+{{\left| {D_y (\Delta\mu^{n})+b_y^{n-1}} \right|}^2} } $. The iterative parameters $b_x$ and $b_y$ are then updated as shown in A-FETV. 

The ADMM-based algorithm for A-FETV and I-FETV is given in Algorithm 1, where ${inner\_loop}$ is the number of iterations for the ADMM-based algorithm. We set ${inner\_loop}$ to 100 in all the experiments in this paper.
\hypertarget{Algorithm 1}{}
\begin{table}[ht] 
\centering
\begin{tabular}{p{12cm}}
\hline
\rowcolor{Gray}
\textbf{Algorithm 1}: ADMM-based algorithm for A-FETV and I-FETV.\\
\hline
\vspace{-5pt}
\;  INPUT: $J$, $y$, ${inner\_loop}$, ${\epsilon_1}$, regularization parameter $\theta > 0$, $\lambda > 0$ \\
\textbf{Initialization}: ${{{\nu}_x^0} ={{\nu}_y^0} = {b_x^0} = {b_y^0} = 0}$  \\
\quad \textbf{for} $n=1:{inner\_loop}$  \\
\quad \textbf \;1: Update $\mu^n$ using Eq. \eqref{eq:Ani4} \\
\quad \textbf \;2: Update ${\nu}_{x}^n$ and ${\nu}_{y}^n$ using Eq. \eqref{eq:Ani6} for A-FETV or Eq. \eqref{eq:Ani13} for I-FETV\\
\quad \textbf \;3: Update $b_{x}^n$ and $b_{y}^n$ using Eq. \eqref{eq:Ani8} \\
\quad \textbf \;4: Stop if $n={{inner\_loop}}$ or $|| {\Delta\mu}^n -{\Delta\mu}^{n-1} ||_{1} / ||{\Delta\mu}^{n-1} ||_{1}\leqslant {\epsilon_1}$, otherwise go to Step 1. \\
\quad \textbf{end for} \\
\textbf{RETURN} ${\Delta\mu}^k = {\Delta\mu}^n$ \\
\hline
\end{tabular} 
\end{table}

We then propose ADMM-based algorithm to address the minimizations of A-GTV and I-GTV. For A-GTV, we first introduce an auxiliary splitting vector variable ${\nu}$, an iterative parameter $b$, and a positive penalty parameter $\theta$. The sizes of ${\nu}$ and $b$ are both of $N \times N$ where $N$ represents the number of vertices. The A-GTV problem can be reformulated as the following unconstrained optimization problem
\begin{equation} \label{eq:graphunconstrain}
{{\Delta\mu}^n,{\nu}^n}  = \arg \mathop {\min }\limits_{{\Delta\mu}, {\nu}}  \left\{ \frac{1}{2} \| {{J}{\Delta\mu}  - {\Delta\Phi}\|_2^2 + \lambda \sum\limits_{i = 1}^N { \| {\nu}_i\|_1 }  + \frac{\theta }{2}\sum\limits_{i = 1}^N {{{\| {{{\nu}_i} - {\nabla _w}(\Delta\mu_i) - {b_i^{n-1}}} \|_2^2}}} } \right\}.
\end{equation}
Since Eq. \eqref{eq:graphunconstrain} is a multivariate minimization problem, we first solve the subproblem with respect to $\Delta\mu$ 
\begin{equation} \label{eq:graphunconstrain1}
{\Delta\mu^n}  = \arg \mathop {\min }\limits_{\Delta\mu}  \left\{ \frac{1}{2} \| {{J}{\Delta\mu}  - {\Delta\Phi}\|_2^2 + \frac{\theta }{2}\sum\limits_{i = 1}^N {{{\| {{{\nu}_i^{n-1}} - {\nabla _w}(\Delta\mu_i) - {b_i^{n-1}}} \|_2^2}}} } \right\},
\end{equation}
which gives the the optimality condition
\begin{equation} \label{eq:graphOpitmal}
{\left( {{J^T}{J}\Delta\mu  - {J^T}{\Delta\Phi}} \right)_i} + \theta {\rm{div}}_w\left( {{{\nu}^{n-1} _i} - {\nabla _w}(\Delta\mu_i) - {b_i^{n-1}}} \right) = 0, \; \; i={1,...,N.}
\end{equation}
With the definition of the \textit{nonlocal divergence operator} (Eq. \eqref{eq:nonlocaldivergence}) and the \textit{nonlocal Laplace operator} (Eq. \eqref{eq:nonlocallap}), the point-wise equation system (Eq. \eqref{eq:graphOpitmal}) can be equivalently converted to the following matrix-based equation system
\begin{equation} \label{eq:graphOpitmal-pw}
({J^{T}}{J} - {\theta L})\Delta\mu  = {{J}^{T}}{\Delta\Phi} - \theta g^{n-1}.
\end{equation}
$L$ above is the graph Laplacian in matrix form, whose entries are 
\[{L_{i,j}} = \left\{ \begin{array}{ll}
-\sum\limits_{j \in {{\cal {N}}_i}}^{} {{w_{ij}}}  & {\rm{if}}\; i=j \\
 \;\;\;\;\;\;\;\;\;\;\;{w_{ij}} & {\rm{otherwise}}
\end{array} \right..\]
In Eq. \eqref{eq:graphOpitmal-pw}, the vector $g^{n-1}=\sum\limits_{j\in { {\cal {N}}_i}}^{} \sqrt{{w_{ij}}} \left( {{\nu}_{ji}^{n-1} - {\nu}_{ij}^{n-1}} \right) +  \sum\limits_{j\in { {\cal {N}}_i}} \sqrt{{w_{ij}}} \left( {b_{ji}^{n-1} - b_{ij}^{n-1}} \right)$. Eq. \eqref{eq:graphOpitmal-pw} is a system of linear equations, the solution $\Delta\mu^n$ can be acquired iteratively using the same method in A-FETV. Then we minimize the following subproblem with respect to ${\nu}$
\begin{equation} \label{eq:equ0}
{\nu}^n = \arg \mathop {\min }\limits_{{\nu}} \left\{ {\lambda \sum\limits_{i = 1}^N { \| {\nu}_i\|_1 }  + \frac{\theta }{2}\sum\limits_{i = 1}^N {{{\| {{{\nu}_i} - {\nabla _w}(\Delta\mu_i^n) - {b_i^{n-1}}} \|_2^2}}} } \right\},
\end{equation}
which has an analytical solution, calculated from the generalized shrinkage formula
\begin{equation} \label{eq:equ1}
{\nu}_{ij}^{n} = {\rm{ max}} \left( {\left| {\sqrt {{w_{ij}}} \left( {\Delta\mu_j^{n} - \Delta\mu_i^{n}} \right) + b_{ij}^{n-1}} \right| - \frac{\lambda }{\theta },0} \right)\frac{{\sqrt {{w_{ij}}} \left( {\Delta\mu_j^{n} - \Delta\mu_i^{n}} \right) + b_{ij}^{n-1}}}{{\left| {\sqrt {{w_{ij}}} \left( {\Delta\mu_j^{n} - \Delta\mu_i^{n}} \right) + b_{ij}^{n-1 }} \right|}},
\end{equation}
with the convention that $0/0=0$. Lastly, we update the iterative parameter $b$ with
\begin{equation} \label{eq:equ2}
b_{ij}^{n} = b_{ij}^{n-1} + \sqrt {{w_{ij}}} \left( {\Delta\mu_j^{n} - \Delta\mu_i^{n}} \right) - {\nu}_{ij}^n.
\end{equation}

We can similarly apply ADMM to the minimization of I-GTV, which can be transformed into the following unconstrained problem with the auxiliary splitting vector variable ${\nu}$, an augmented Lagrangian multiplier $b$, and a positive penalty parameter $\theta$.
\begin{equation} \label{eq:equ3}
{\Delta\mu^n,{\nu}^n}  = \arg \mathop {\min }\limits_{\Delta\mu, {\nu}}  \left\{ \frac{1}{2} \| {{J}{\Delta\mu}  - {\Delta\Phi}\|_2^2 + \lambda \sum\limits_{i = 1}^N { \| {\nu}_i\|_2 }  + \frac{\theta }{2}\sum\limits_{i = 1}^N {{{\| {{{\nu}_i} - {\nabla _w}(\Delta\mu_i) - {b_i^{n-1}}} \|_2^2}}} } \right\}.
\end{equation}
The L$_2$ subproblem with respect to $\Delta\mu$ is the same as the one in A-GTV and can be computed with Eq. \eqref{eq:graphOpitmal-pw}. We then fix $\Delta\mu$ to minimize the second subproblem with respect to ${\nu}$:
\begin{equation} \label{eq:equ4}
{\nu}^n  = \arg \mathop {\min }\limits_{{\nu}}  \left\{ {\lambda \sum\limits_{i = 1}^N \| {\nu}_i\|_2  + \frac{\theta }{2}\sum\limits_{i = 1}^N {{{\| {{{\nu}_i} - {\nabla _w}(\Delta\mu_i^n) - {b_i^{n-1}}} \|_2^2}}} } \right\},
\end{equation}
which can be solved with the following soft thresholding equation
\begin{equation} \label{eq:equ5}
{\nu}_{ij}^{n} = {\rm{ max}}\left( {\sqrt {\sum\limits_{j \in {{\cal {N}}_i}}^{} {{{\left( {\sqrt {{w_{ij}}} \left( {\Delta\mu_j^{n} - \Delta\mu_i^{n}} \right) + b_{ij}^{n-1}} \right)}^2}} }  - \frac{\lambda }{\theta } ,0} \right)\frac{{\sqrt {{w_{ij}}} \left( {\Delta\mu_j^{n} - \Delta\mu_i^{n}} \right) + b_{ij}^{n-1}}}{{\sqrt {\sum\limits_{j \in {{\cal {N}}_i}}^{} {{{\left( {\sqrt {{w_{ij}}} \left( {\Delta\mu_j^{n} - \Delta\mu_i^{n}} \right) + b_{ij}^{n -1}} \right)}^2}} } }},
\end{equation}
with the convention that $0/0=0$. The update of iterative parameter $b$ is the same as for A-GTV, as shown in Eq. \eqref{eq:equ2}. The ADMM-based algorithm for A-GTV and I-GTV is given in Algorithm 2.
\hypertarget{Algorithm 2}{}
\begin{table}[ht] 
\centering
\begin{tabular}{p{12cm}}
\hline
\rowcolor{Gray}
\textbf{Algorithm 2}: ADMM-based algorithm for I-GTV and A-GTV.\\
\hline
\vspace{-5pt}
\;  \textbf{INPUT}: ${\rm{J}}$, $y$, ${inner_{-}loop}$, ${\epsilon_1}$, regularization parameter $\theta > 0$, $\lambda > 0$  \\
\textbf{Initialization}: ${{{\nu}^0} = {b^0} = 0}$  \\
\quad \textbf{for} $n=1:{inner_{-}loop}$  \\
\quad \textbf \;1: Update $\Delta\mu^n$ using Eq. \eqref{eq:graphOpitmal-pw} \\
\quad \textbf \;2: Update ${\nu}^n$ using Eq. \eqref{eq:equ1} for A-GTV or Eq. \eqref{eq:equ5} for I-GTV \\
\quad \textbf \;3: Update $b^n$ using Eq. \eqref{eq:equ2} \\
\quad \textbf \;4: Stop if $n={inner_{-}loop}$ or $|| \Delta\mu^n -\Delta\mu^{n-1} ||_{1} / ||\Delta\mu^{n-1} ||_{1}\leqslant {\epsilon_1}$, otherwise go to Step 1. \\
\textbf{RETURN} $\Delta\mu^k = \Delta\mu^n$ \\
\hline
\end{tabular} 
\end{table}

Therefore the whole procedure for minimizing the TV-regularized inverse problem (Eq. \eqref{eq:twostepLinears}) is given in Algorithm 3, in which ${outer_{-}loop}$ represents the number of iterations required for the DOT reconstruction and is set to 40 for all experiments in this paper. 
\hypertarget{Algorithm 3}{}
\begin{table}[ht] 
\centering
\begin{tabular}{p{12cm}}
\hline
\rowcolor{Gray}
\textbf{Algorithm 3}: Algorithm for minimizing the TV-associated inverse problem.\\
\hline
\vspace{-5pt}
\; \textbf{INPUT}: $\Phi ^{\rm{M}}$, ${\cal F}\left(\cdot \right)$, $\mu^0$, ${outer_{-}loop}$, ${\epsilon_2}$  \\
\quad \textbf{for} $k=1:{outer_{-}loop}$  \\
\quad \textbf \;1: Compute ${\cal F}\left( \mu^{k-1} \right)$ and ${J^{k - 1}}$ \\
\quad \textbf \;2: Set $\Delta \Phi^{k-1} = {\Phi ^{\rm{M}}} - {\cal F}\left( {{\mu ^{k - 1}}} \right)$ \\
\quad \textbf \;3: Compute $\Delta\mu^k$ by introducing $\Delta \Phi^{k-1}$ and ${J^{k - 1}}$ to one of \textbf{Algorithm 1-2} \\
\quad \textbf \;4: Update $\mu^k = \Delta\mu^k + \mu^{k-1}$ \\
\quad \textbf \;4: Stop if $k={outer_{-}loop}$ or $(|| {\cal F}\left( \mu^k \right)-\Phi ^{\rm{M}} ||_2^2 - || {\cal F}\left( \mu^{k-1} \right)-\Phi ^{\rm{M}} ||_2^2) / || {\cal F}\left( \mu^{k-1} \right)-\Phi ^{\rm{M}} ||_2^2 \leqslant {\epsilon_2}$, otherwise go to Step 1. \\
\quad \textbf{end for} \\
\textbf{RETURN} $\mu^k$ \\
\hline
\end{tabular} 
\end{table}

\section{Experiments}
\begin{figure}[ht]
\centering
\includegraphics[width=0.8\textwidth]{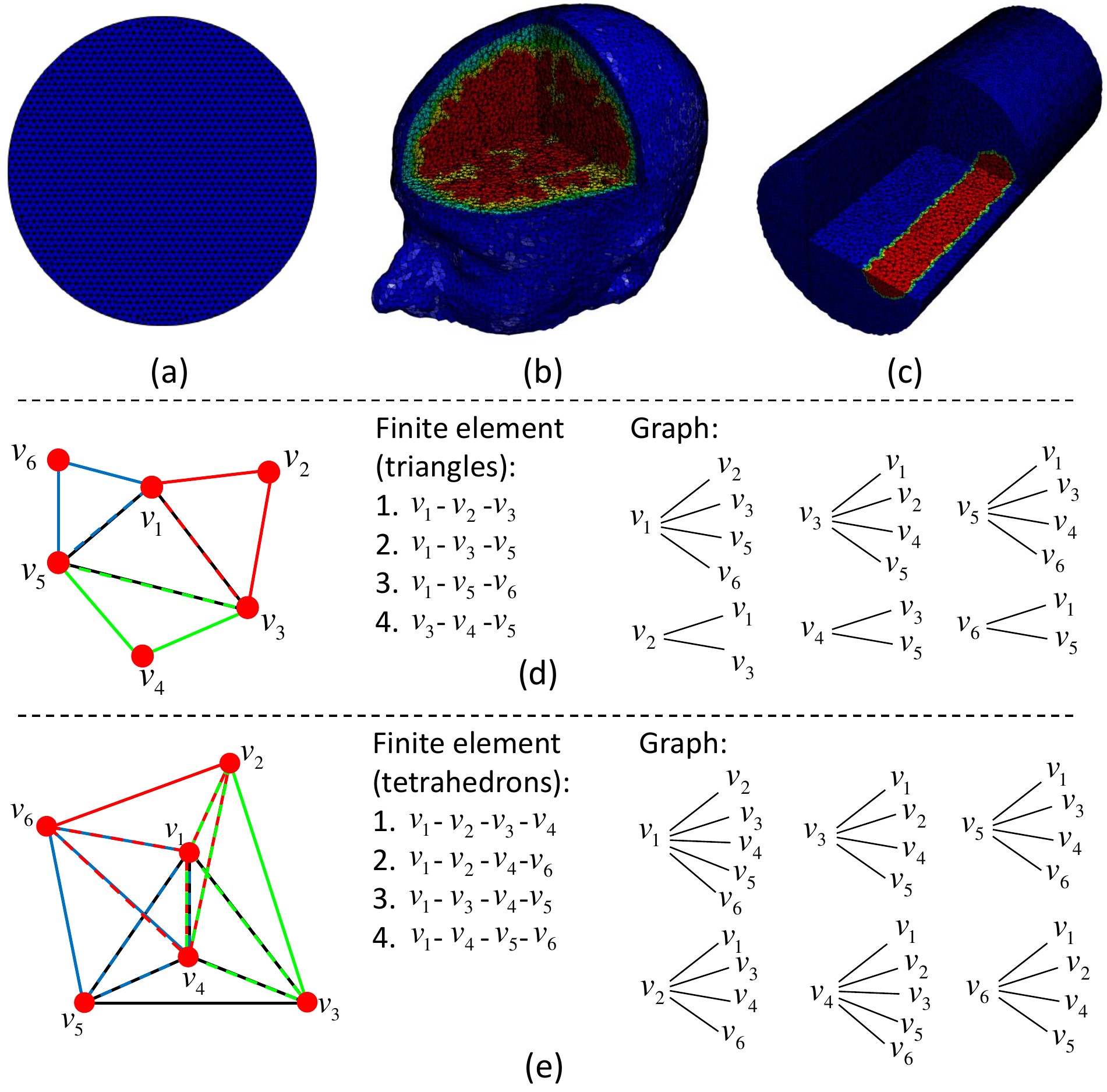}
\caption{(a)-(c): Discretized computational domain of the three experimental samples; (d): Detailed mesh composition of 2D geometry in finite element and graph representation respectively; (e): Detailed mesh composition of 3D geometry in finite element and graph representation respectively.}
\label{fig:f0}
\end{figure}

In this section, we describe extensive experiments to qualitatively and quantitatively evaluate the performance of finite element and graph representations on the two variants of TV regularization in DOT. We use the finite element representation for the forward modelling in all the experiments and use both the finite element and graph representations to discretize the TV regularization term during the solution of the inverse problem. We first define four evaluation metrics to quantify the quality of the reconstructed images. Then we describe simulated numerical experiments on 2D circle and 3D head samples, and real experiments performed on phantom samples. Fig.~{\ref{fig:f0}} shows the unstructured grids of the three computational domains. Red dots represent the vertices in the computational domain. In 2D, using the finite element representation, the computational domain is discretized with a finite number of triangles (Fig.~{\ref{fig:f0}} (d)) while in 3D, tetrahedra are taken as the basic element  (Fig.~{\ref{fig:f0}} (e)). However the graph representation is the same in both 2D and 3D because the graph method requires only vertices and edges of the mesh. 
For simulated experiments in which measurement noise was added, ten repeats were performed. In all experiments, the forward model was implemented using the NIRFAST package \cite{dehghani2009near} in Matlab R2017a (Mathworks, Natick, USA). The simulated experiments conducted are all based on  single wavelength continuous-wave (CW) measurements where the optical property to be recovered is the tissue absorption coefficient $\mu_a$ at that wavelength.

\subsection{Quantitative Evaluation Metrics}
The quantitative evaluation is performed using four evaluation metrics: the localization error, average contrast, peak signal-to-noise ratio (PSNR) and relative recovered volume. If the reconstructed image is identical to the ground truth image, the localization error is 0, average contrast and relative recovered volume are both equal to 1. PSNR is higher if the reconstructed image is closer to the ground truth image. 

Localization error is defined as the Euclidian distance between the central nodes $\rm{X_s}$ of the simulated activation region and $\rm{X_r}$ of the recovered activation region. The recovered activation regions are selected by thresholding the recovered changes based on 60\% of the maximum recovered changes.
\begin{equation}
\mbox{Localization error} = {\left\| {{X_s} - {X_r}} \right\|_2}.
\label{eq:LE}
\end{equation}

The second evaluation metric is the average contrast which is based on the mean value of the region of interest:
\begin{equation}
\mbox{Average contrast} =  \left ( {\sum\limits_{i = 1}^{N_r} {\mu ^i} /{N_r}} \right )/{\widetilde {\mu}},
\label{eq:AC}
\end{equation}
where $\mu ^i$ denotes the recovered optical property on the finite element node $i$. ${N_r}$ is the number of nodes in the recovered activation region. ${\widetilde \mu}$ is the ground truth values of the optical property in the recovered activation region. 

PSNR is the third evaluation metric, which aims to evaluate the difference between the ground truth image and the recovered image. Larger PSNR values means less difference between these two types of images. PSNR is defined as follows
\begin{equation}
\mathrm{PSNR} = 10 \cdot {\log _{10}}\left( {\mathrm{MAX}_{{\mu}}^2} / {\mathrm{MSE}}\right).
\label{eq:psnr}
\end{equation}
Here, ${\mathrm{MAX}_{{\mu}}}$ is the maximum pixel value of $\mu$ and $\mathrm{MSE}$ is the mean squared error between the recovered and ground truth images with $\mathrm{MSE} = {\sum_{i = 1}^{N}} {{{\left( {\mu^i - \tilde \mu^i} \right)}^2}}/N $.

Finally, we measure the relative recovered volume which is given as
\begin{equation}
{{\rm{V}}_{RRV}} = {{\rm{V}}_r}/{{\rm{V}}_s} \times 100\%  ,
\label{eq:RV}
\end{equation}
where ${{\rm{V}}_r}$ and ${{\rm{V}}_s}$ denote the volume of the recovered activation region and simulated activation region, respectively.

\subsection{Experiments on anisotropic TV regularization}
Anisotropic TV regularization is easy to implement because the partial derivatives along different directions can be decoupled as explained in Section 3.2. It is based on the assumption that the shape of the region of interest is aligned with the coordinate axes. Its minimization favors horizontal and vertical structures, because oblique structures cause the TV regularization to increase \cite{condat2017discrete}. In DOT, this assumption does not necessarily hold as the region of interest is normally random and structures are not normally aligned with the coordinate system. Therefore anisotropic TV regularization seems to be a poor choice for discrete TV in DOT, as it yields 'blocky' artefacts. However no research has been carried out about the relationship between the 'blocky' artefacts and the representation employed to discretize over the unstructured computational domain. In this section we investigate the anisotropic TV regularization in DOT reconstruction and compare their FE- and graph-based implementations. The effect of the representation method adopted on anisotropic TV regularization will be evaluated.

A two dimensional (2D) circular geometry is simulated with one anomaly centered at (-10mm,10mm). The 2D model has a radius of 43mm while the radius of the anomaly is 10mm. Sixteen source-detector fibres are placed equidistant around the external boundary for CW boundary data acquisition. When one fibre as a source is turned on, the rest are used as detectors, leading to 240 total boundary data points per wavelength. All sources were positioned one scattering distance within the outer boundary because the source is assumed to be spherically isotropic. In order to evaluate the effect of mesh resolution on the representation method, two reconstruction meshes are created with different spatial resolutions. The coarser mesh has 1785 nodes and 3418 linear triangle elements with the average element size 1.6977mm${^2}$ (Fig. \ref{fig:f1} (a)) while the finer one has 5133 nodes and 10013 elements with the average element size 0.5801mm${^2}$ (Fig. \ref{fig:f1} (d)). The background absorption coefficient $\mu_a$ is set as 0.01$\rm{mm^{-1}}$ and $\mu_a$ for the anomaly is set as 0.03$\rm{mm^{-1}}$ (Fig. \ref{fig:f1} (b) and (e)). $\mu_s$ remains constant as 1$\rm{mm^{-1}}$. To represent various realistic cases, normally distributed randomly generated Gaussian noise ranging from 0\% to 3\% at 1\% intervals was added to the boundary measurements. Reconstructed images of the absorption coefficient are shown in Figs. \ref{fig:f1} (c) and (f).

\subsection{Experiments on isotropic TV regularization}
\subsubsection{Two dimensional circular experiments}
Using the same reconstruction meshes described in section 4.2, we compare I-FETV, I-GTV against a baseline Tikhonov model. To represent various realistic cases, normally distributed randomly generated noise ranging from 0\% to 3\% at 1\% intervals was added to the amplitude of the boundary data. Reconstructed images of absorption coefficient are shown in Figs. \ref{fig:f4} (c) and (f). The 1D cross sections and evaluation metrics comparisons are displayed in Fig. \ref{fig:f5} and Fig. \ref{fig:f6}.

\subsubsection{Three dimensional head numerical experiments}
We now evaluate the isotropic TV model with two discrete differential operator definitions on a physically realistic three dimensional head model. This was created from T1-weighted MPRAGE scans acquired by Eggebrecht \textit{et al}~\cite{eggebrecht2014mapping} using the process described by Wu et al \cite{wu2014quantitative} in which Statistical Parametric Mapping (SPM) software\cite{ashburner2003image} was used to perform parametric segmentation of five tissue types (scalp, skull, cerebrospinal fluid (CSF), gray matter, white matter) based on the pixel intensity probability function distribution. The five layers were processed in NIRFAST to create masks and layered volumetric FEM meshes. The reconstruction mesh consists of 50721 nodes associated with 287547 tetrahedral elements, with the average element size 9.2676mm$^{3}$. Each node is labeled by one of the five segmented head tissue types. Absorption coefficients assigned to each layer are from an in vivo study \cite{eggebrecht2012quantitative} at 750nm (Table~\ref{tb:headpropertyforfive}). 

A high-density (HD) imaging array with 158 sources and 166 detectors (Fig.~{\ref{fig:f7}} first column) \cite{eggebrecht2014mapping} was placed over the whole head, with source-detector (SD) separation distances ranging from 1.3 to 4.8cm. In this study, 3478 differential measurements per wavelength were used to image hemodynamic changes in the brain. Two distinct anomalies were simulated simultaneously in the brain, with each 15mm radius. In order to simulate the traumatic brain injury (TBI) cases where tissue oxygen saturation ($\rm{StO_2}$) is normally between 50\% and 75\% \cite{clancy2015comparison,ichai2017metabolic}, the absorption coefficient in the two anomalies are calculated using Beer's law \cite{dehghani2009near} with 55\% $\rm{StO_2}$ (Fig.~{\ref{fig:f7}} second column). In line with the expected in vivo performance of the imaging system, 0.12$\%$, 0.15$\%$, 0.41$\%$ and 1.42$\%$ Gaussian random noise was added to first (13mm), second (30mm), third (40mm) and fourth (48mm) nearest neighbor measurements to provide realistic data \cite{dehghani2009depth}. Reconstructed absorption coefficients using different model are displayed in the third to fifth column of Fig.~{\ref{fig:f7}}. Corresponding 2D slices are displayed in Fig.~{\ref{fig:f9}} and the evaluation metrics are presented in Fig.~{\ref{fig:f10}}. 

\subsubsection{Experiments with phantom data}
In the final experiment we evaluate different methods on real experimental data which is collected from a solid plastic cylindrical phantom using one non-contact CW-DOT system designed for hand imaging \cite{lighter2017multispectral}. The phantom has size of radius 12.3mm and length 50mm. 35 sources and 99 detectors are positioned on the underside and top of the phantom respectively (Fig.~{\ref{fig:f12}}(a)). The absorbing dye within the phantom was treated as a chromophore that has unit concentration in the bulk of the phantom. Its extinction coefficient was modelled by the measured absorption coefficient. A cylindrical rod was placed at the depth of 5mm to simulate the heterogeneous version of the phantom (Fig.~{\ref{fig:f12}}(a)). The rod has radius 3mm and length 50mm and provides a 2:1 contrast in dye concentration compared to background (Fig.~{\ref{fig:f12}} (c) top row). Five wavelengths (650nm, 710nm, 730nm 830nm and 930nm) in a transmission setup are used to collect the boundary data. The reconstruction mesh consists of 9082 nodes and 48099 linear tetrahedral elements with the average tetrahedral elements size 0.4218mm$^{3}$. Ground truth data and images reconstructed with Tikhonov, I-FETV and I-GTV are shown in Fig.~{\ref{fig:f12}}(c). The four evaluation metrics in the volume of illumination are given in Table~\ref{tb:compa1}. For all the experiments above, the regularization parameter $\lambda$ is determined using an L-curve method\cite{lu20181}.

\section{Discussion}

\begin{figure}[ht]
\centering
\includegraphics[width=0.8\textwidth]{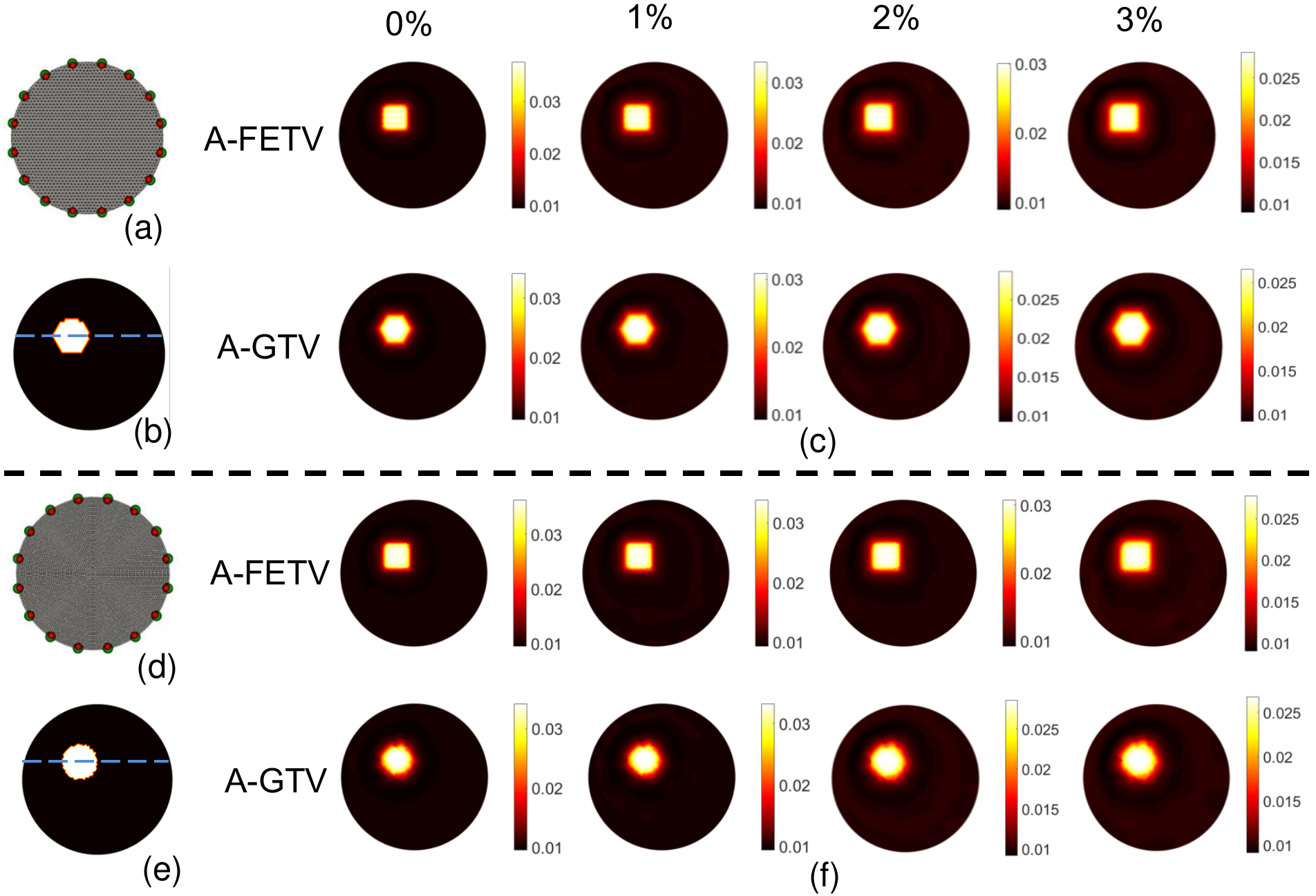}
\caption{(a)-(c): Reconstruction on the 2D mesh with low spatial resolution. (d)-(f): Reconstruction on the 2D mesh with high spatial resolution. (a) and (d): 2D reconstruction mesh with sixteen co-located sources and detectors. (b) and (e) give the original target distributions. First row in (c) and (f) represents the results using A-FETV on 0\% , 1\% , 2\% and 3\% noisy data while the second row shows the results using A-GTV.}
\label{fig:f1}
\end{figure}

The 2D images reconstructed using A-FETV and A-GTV are shown in Fig.~{\ref{fig:f1}}, together with the original target distributions. The results show that A-FETV keeps reconstructing the target with boundaries that align with the coordinate axes which is same to the assumption of the anisotropic TV regularization. In addition, some artefacts are observed inside the recovered region of interest when the mesh resolution is low. The reconstructions using A-GTV do not feature these artefacts, and the recovered shape is more accurate, with no bias towards the coordinate axes. The experiments reveal that the blocky artefacts of anisotropic TV regularization are associated with the discretization method used. The blocky artefacts are clearly visible in reconstructions based on the finite element representation, but not in the ones based on the graph representation. This is because in the graph representation, the region is discretized along all edge-based directions, leading to nearly isotropic solutions. Therefore in DOT, A-GTV can adapt to the ground truth solution. However, it is not a good method to preserve anisotropy if anisotropy is a desired property of the solution.

\begin{figure}[ht]
\centering
\includegraphics[width=0.7\textwidth]{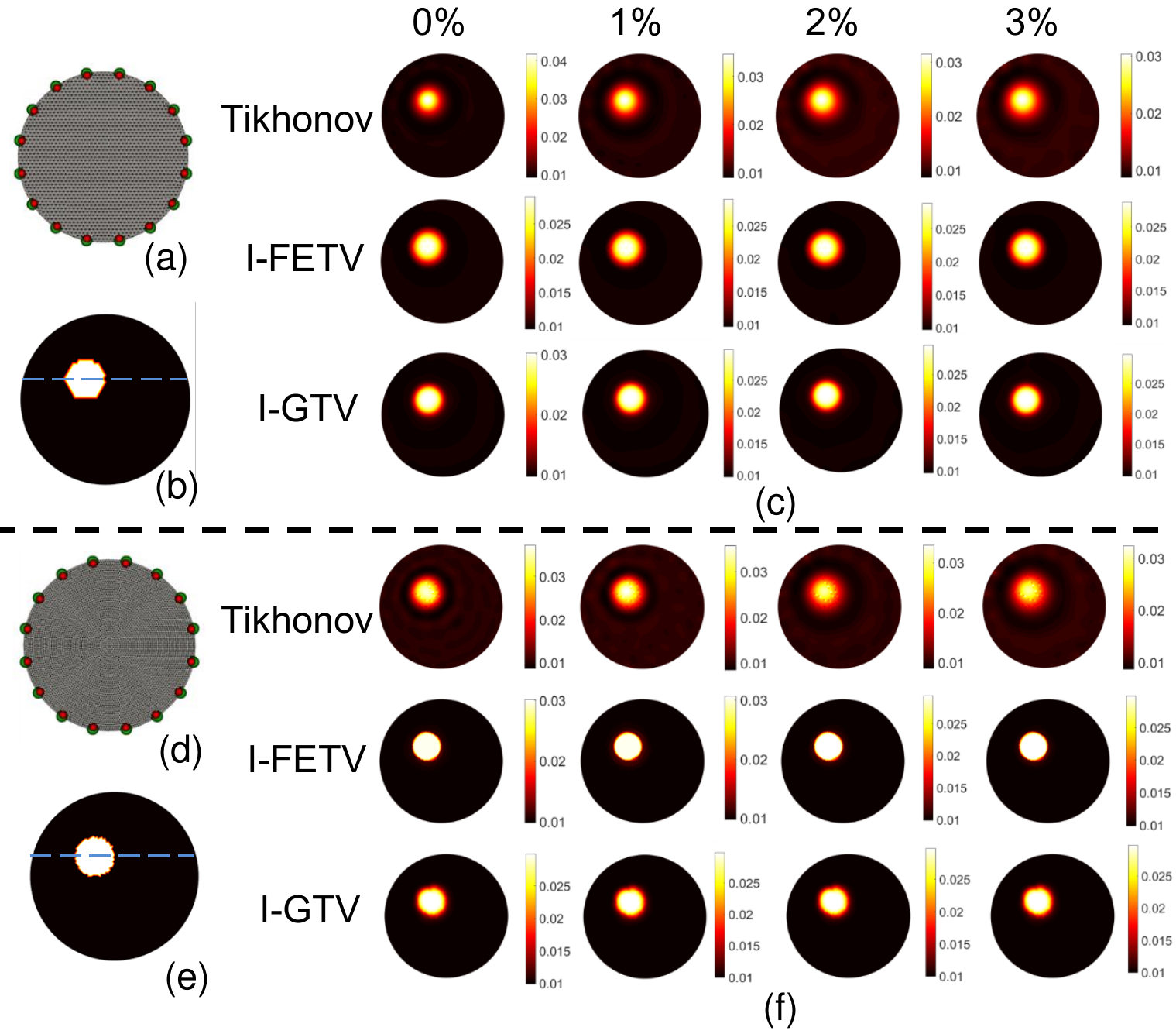}
\caption{(a)-(c): Reconstruction on the 2D mesh with low spatial resolution; (d)-(f): Reconstruction on the 2D mesh with high spatial resolution. (a) and (d): 2D reconstruction mesh with sixteen co-located sources and detectors. (b) and (e) give the original target distributions. First row in (c)and (f) represents the results using I-FETV on 0\% , 1\% , 2\% and 3\% noisy data while the second row shows the results by I-GTV.}
\label{fig:f4}
\end{figure}

\begin{figure}[ht]
\centering
\includegraphics[width=1\textwidth]{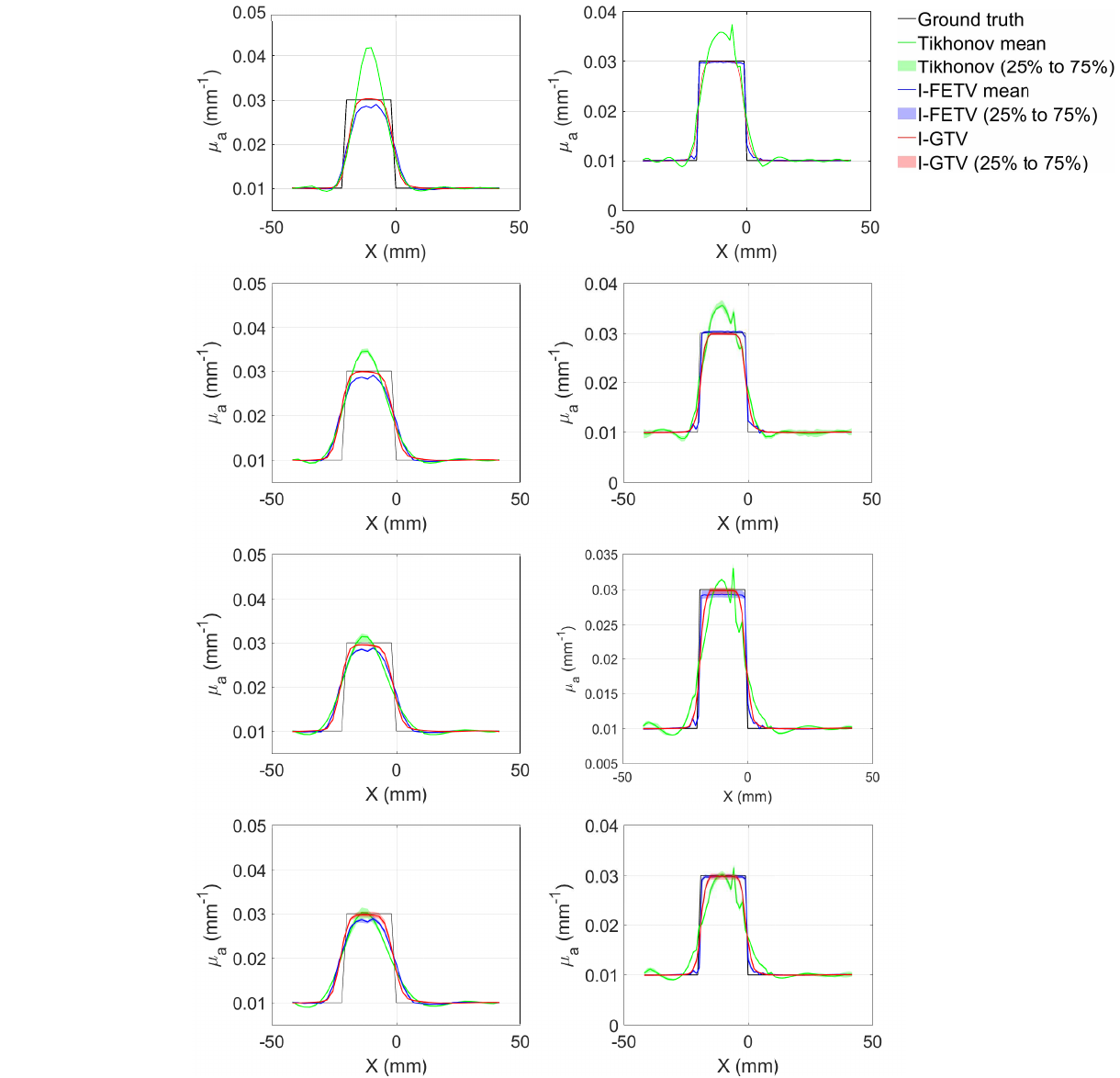}
\caption{1D cross section of images recovered in Fig.~{\ref{fig:f4}}. First column corresponds to Fig.~{\ref{fig:f4}} (c) where the spatial resolution of the reconstruction mesh is lower. Second column corresponds to Fig.~{\ref{fig:f4}} (f) where the spatial resolution of the reconstruction mesh is higher. Top to bottom row: 0\%, 1\%, 2\% and 3\% added Gaussian noise.}
\label{fig:f5}
\end{figure}

\begin{figure}[ht]
\centering
\includegraphics[width=0.85\textwidth]{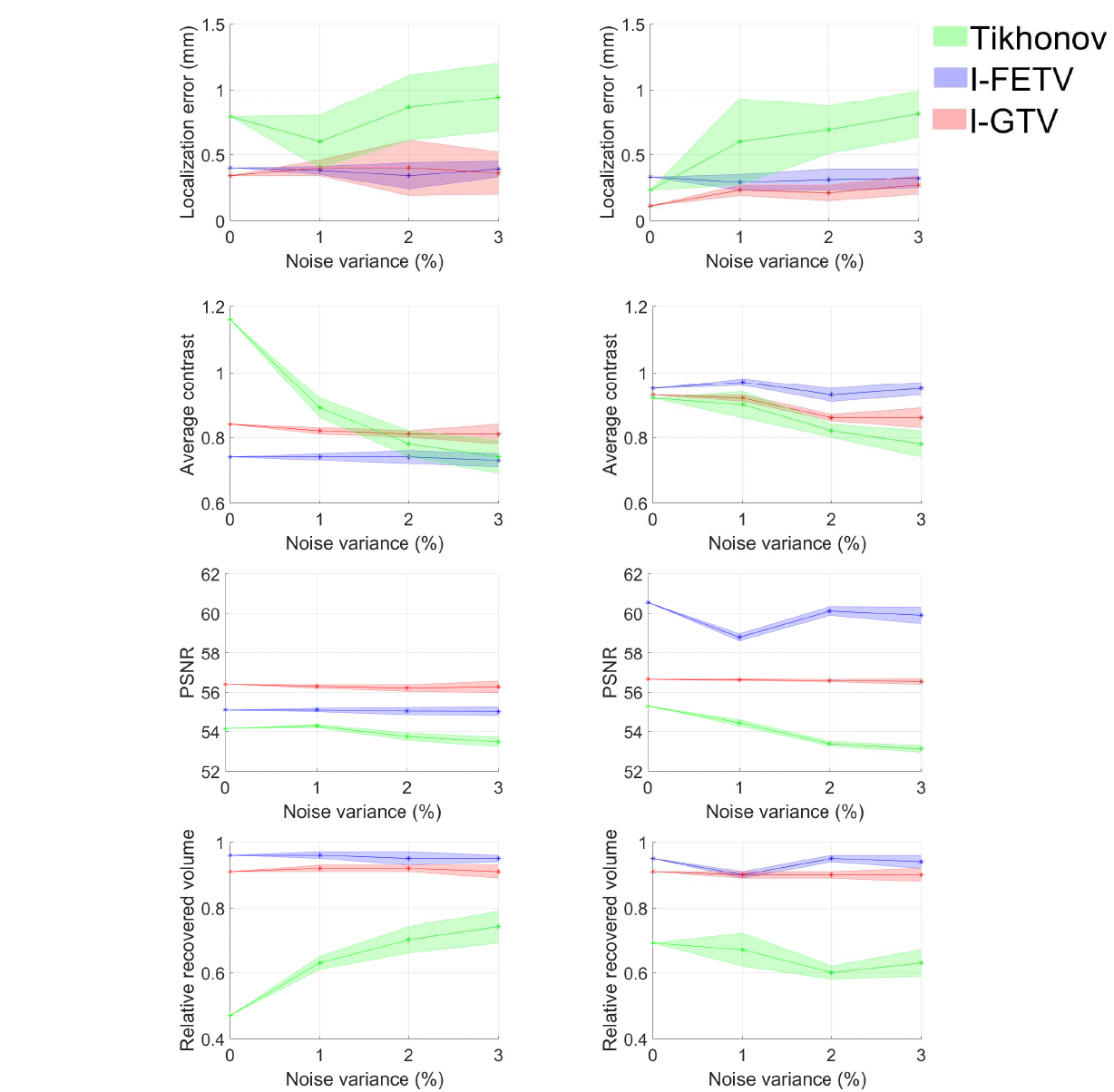}
\caption{Evaluation metrics comparing the performance of different methods at four different noise levels. Top to bottom row: localization error index; average contrast index; PSNR index and relative recovered volume. Left column corresponds to the reconstructions in Fig.~{\ref{fig:f4}} (c) where the reconsturction mesh resolution is low. Right column corresponds to Fig.~{\ref{fig:f4}} (f) where the reconsturction mesh resolution is relatively high.}
\label{fig:f6}
\end{figure}

For the 2D case which uses isotropic TV regularization, Fig.~{\ref{fig:f4}}, Tikhonov reconstruction over-smooths the results and smears the edges. The results become smoother with increases in measurement noise. Little difference can be visually observed between the reconstruction by I-FETV and I-GTV when the reconstruction mesh resolution is low. However when the reconstruction mesh has higher resolution, Fig.~{\ref{fig:f4}} (f), the results by I-FETV is visually closer to the ground truth than the ones by I-GTV. Similar findings are observed from the corresponding 1D cross sections (Fig.~{\ref{fig:f5}}). Tikhonov reconstruction produces a single peaked distribution in the piecewise constant target area, and edges of the objects are over-smoothed. Both TV methods are able to reconstruct a piecewise constant distribution. However when the mesh resolution is lower (first column in Fig.~{\ref{fig:f5}}), fluctuations in the target regions are observed in the results by I-FETV. When the mesh resolution is higher (second column in Fig.~\ref{fig:f5}), the cross-section from I-FETV reconstruction is almost identical to the ground truth. In Fig.~{\ref{fig:f6}}, red and blue areas represent 25$\%$ to 75$\%$ value among the ten repeats' experiments. We see that the performance of I-FETV improves with an increase of mesh resolution: by 25\% in localization error, 26\% in average contrast and 11\% in PSNR, while the performance of I-GTV is relatively unaffected  by the mesh resolution. These 2D experiments confirm that the discrete differential operators based on graph representation are not affected by the mesh resolution while the ones based on a finite element representation become more accurate when the reconstruction mesh is finer.

\begin{figure}[ht]
\centering
\includegraphics[width=0.9\textwidth]{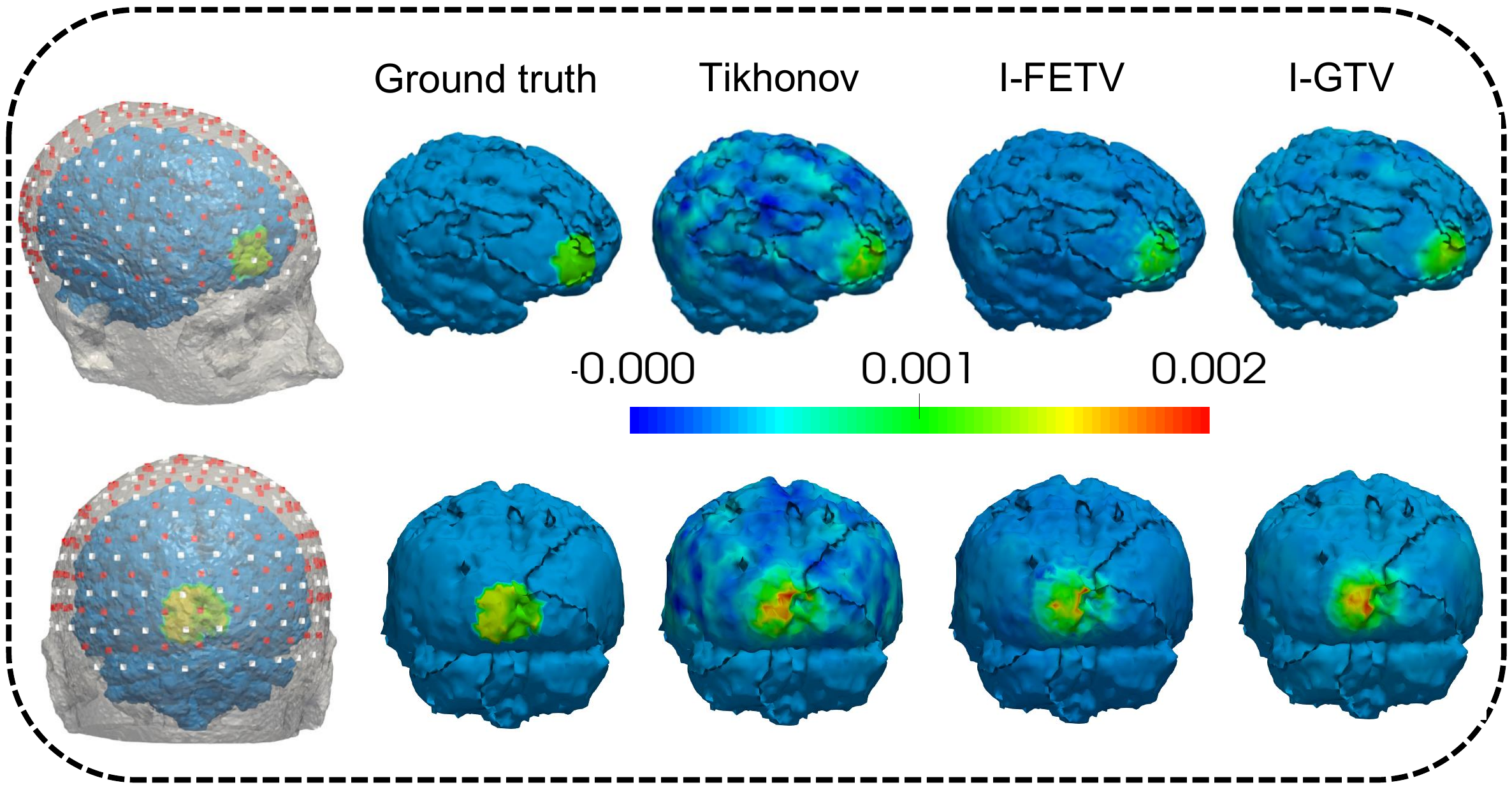}
\caption{First column: distribution of the imaging array with 158 sources (red dots) and 166 detectors (white dots) and the positions of the two simultaneous simulated anomalies. Second to final column: Ground truth and reconstructions by Tikhonov, I-FETV and I-GTV.}
\label{fig:f7}
\end{figure}

\begin{figure}[ht]
\centering
\includegraphics[width=0.9\textwidth]{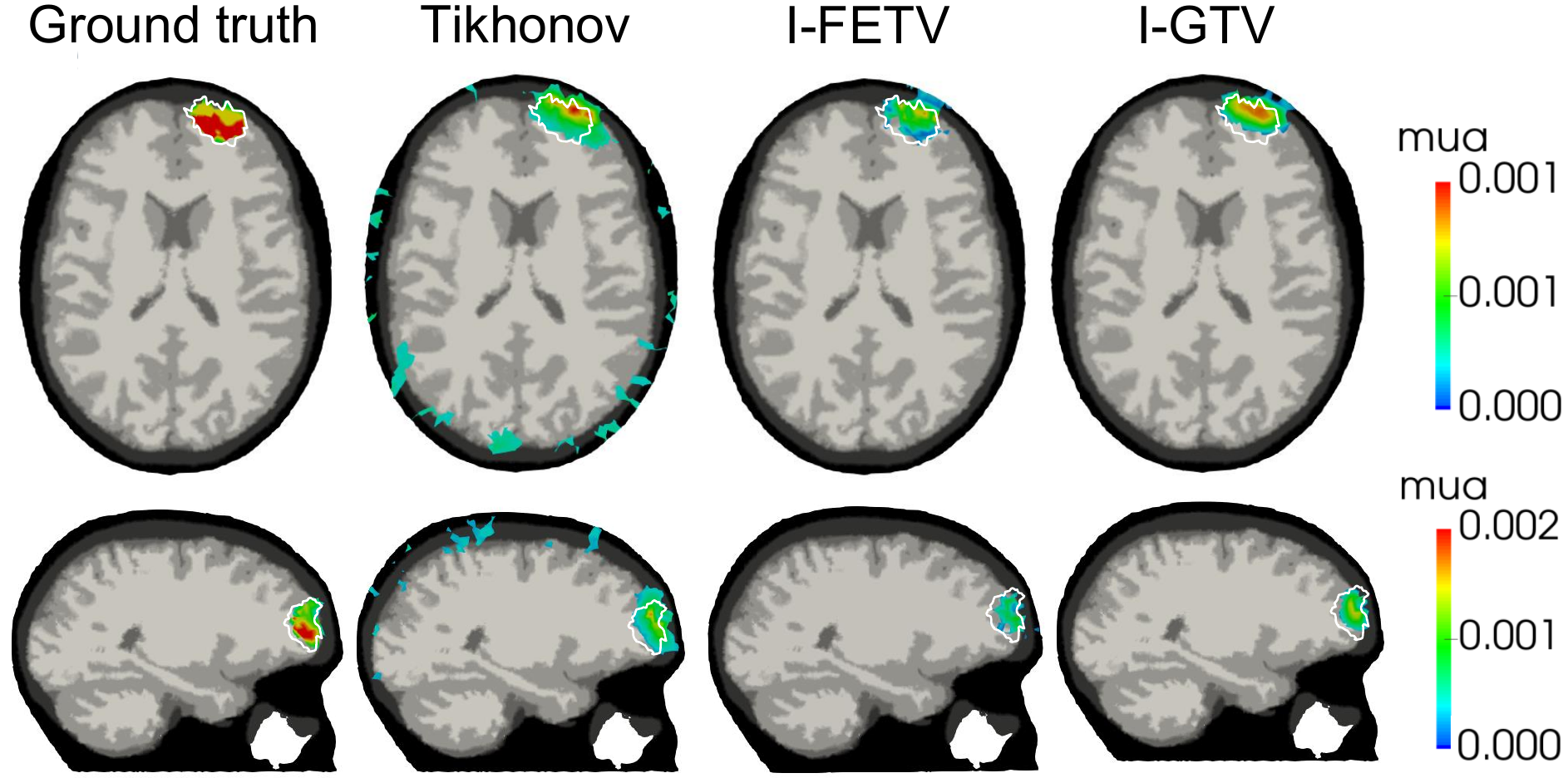}
\caption{2D slices of the reconstructions of the absorption coefficient changes on the forehead anomaly (first row in Fig.~{\ref{fig:f7}}). The ground truth areas are highlighted in white ellipses.}
\label{fig:f9}
\end{figure}

\begin{figure}[ht]
\centering
\includegraphics[width=0.9\textwidth]{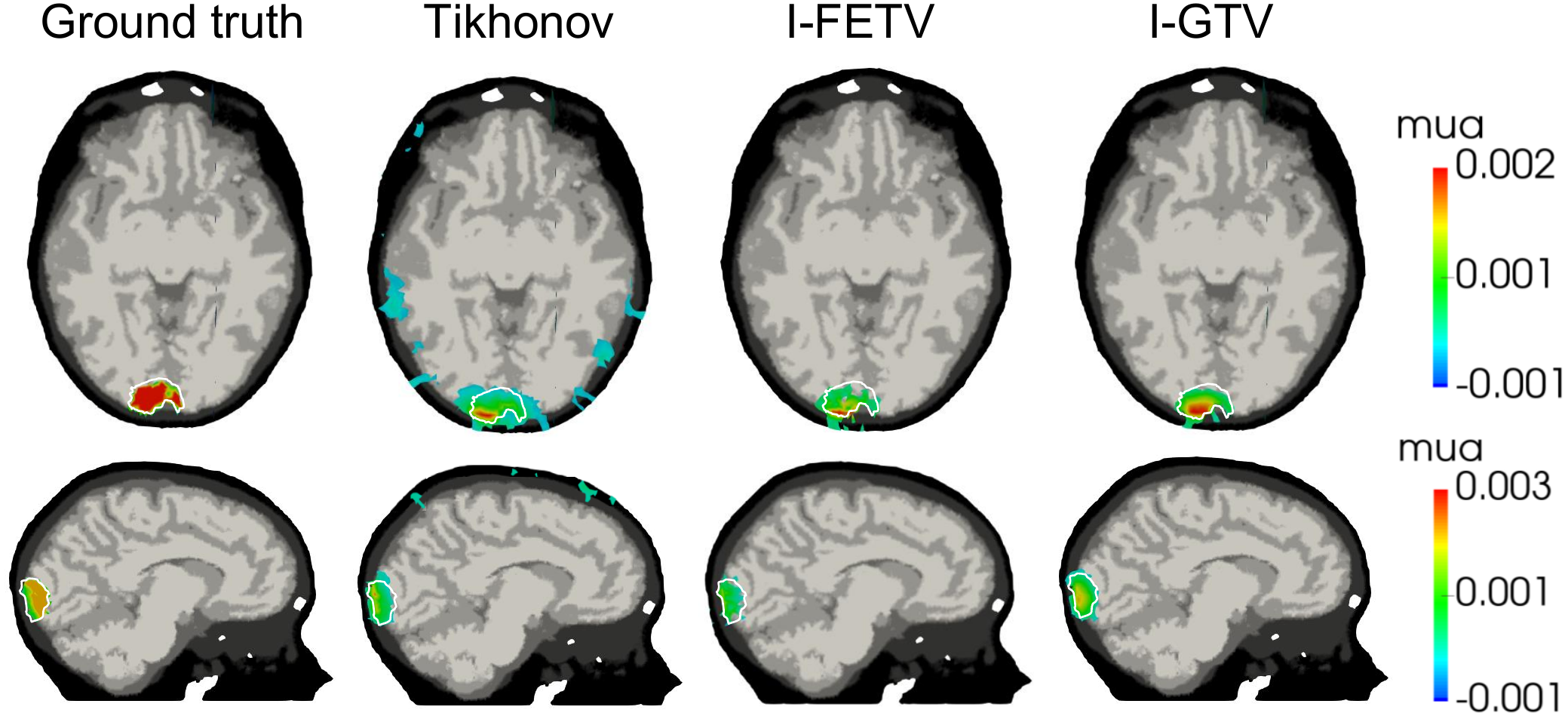}
\caption{2D slices of the reconstructions of the absorption coefficient changes on the back-head anomaly (second row in Fig.~{\ref{fig:f7}}). The ground truth areas are highlighted in white ellipses.}
\label{fig:f10}
\end{figure}

\begin{figure}[ht]
\centering
\includegraphics[width=0.7\textwidth]{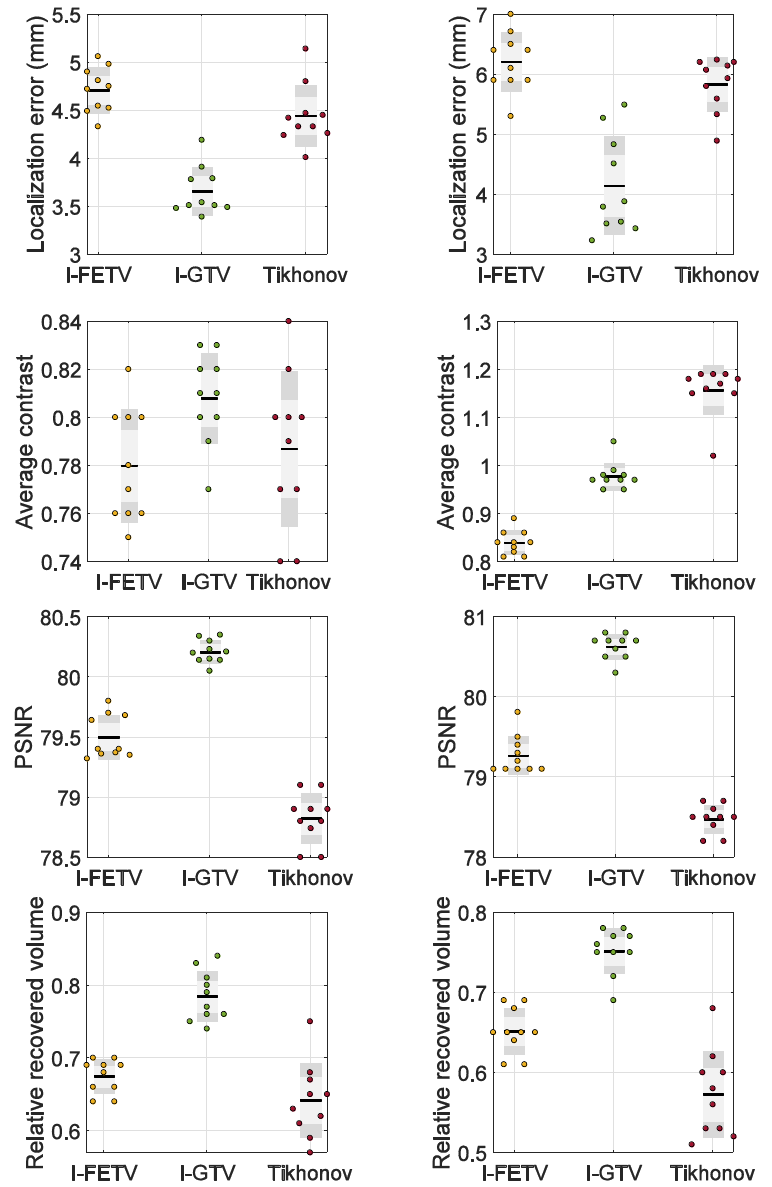}
\caption{Evaluation metrics comparing the performance of different methods on a 3D head model. The left column represents the reconstruction of the forehead anomaly (first row in Fig.~{\ref{fig:f7}}), while the right column gives the reconstruction of the back-head anomaly (second row in Fig.~{\ref{fig:f7}}).}
\label{fig:f11}
\end{figure}

The 3D images reconstructed from the head geometry represent a physically realistic case, in which two anomalies are simulated simultaneously in the brain. From Fig.~{\ref{fig:f7}}, Tikhonov reconstruction lead to many visible artefacts near the source and detector area. Due to smoothing induced by Tikhonov regularization, sharp features are not present in the image recovered. I-FETV and I-GTV both can eliminate the surface artefacts resulting from Tikhonov regularization and reconstruct tightly localized results. These findings can be clearly observed in the 2D slice images shown in Fig.~{\ref{fig:f9}} and Fig.~{\ref{fig:f10}}. It should be noticed that, in Fig.~{\ref{fig:f9}}, the colorbar values corresponding to the green and red parts remain 0.001. It is because only three digits are selected after the radix point and in this study we use rounding off to constrain the three digits. From the visualization of the results, there is no obvious difference between the reconstruction performance of I-FETV and I-GTV because both are based on TV regularization. However it can be observed from the evaluation metrics comparison in Fig.~{\ref{fig:f11}} that I-GTV achieves the lowest localization error, highest peak signal-to-noise ratio and average contrast much closer to 1. The average relative recovered volume achieved by I-GTV is 77\%, compared with I-FETV (66\%) and Tikhonov (64\%). This experiment confirms the lower performance of I-FETV on reconstruction meshes with low spatial resolution.

\begin{table}[ht]
\caption{Head tissue optical property for each of five layers.}
\centering 
\label{tb:headpropertyforfive}
{
  \begin{tabular}{|c||c|c|c|c|c|}
\hline
&Scalp   & Skull  & CSF  & Gray Matter  & White Matter    \\ \hline 
{${\mu _a}$ (${\rm{m}}{{\rm{m}}^{ - 1}}$)} & 0.017      & 0.012 & 0.004 & 0.018 &0.017  \\ \hline
{${\mu _s^\prime}$ (${\rm{m}}{{\rm{m}}^{ - 1}}$)} & 0.74      & 0.94 & 0.3 & 0.84  &1.19  \\ \hline            
  \end{tabular}
  } 
\end{table}

\begin{figure}[ht]
\centering
\includegraphics[width=0.99\textwidth]{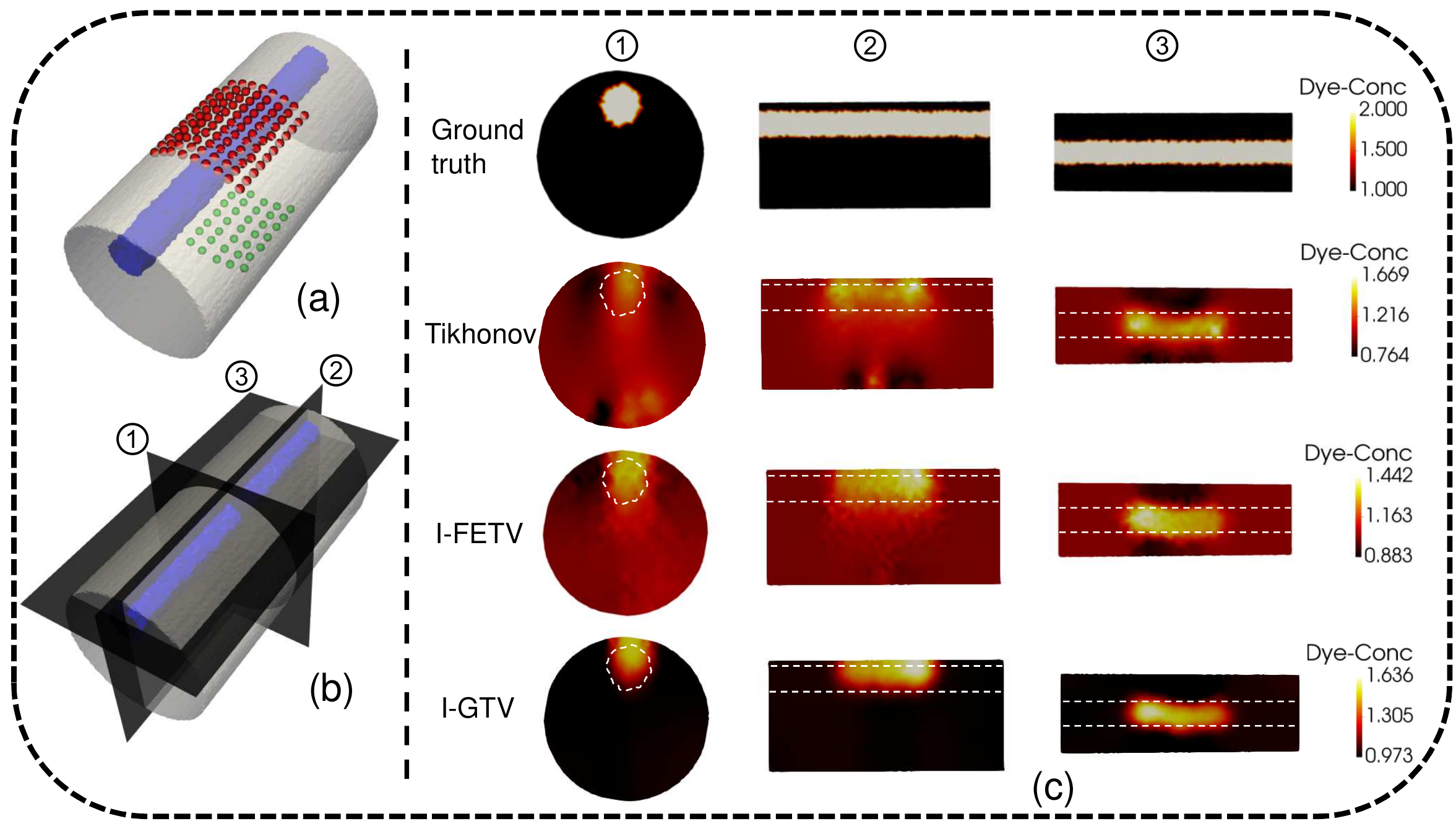}
\caption{(a): Distribution of sources and detectors. (b): Illustration of the overall distribution of three slices. (c): Ground truth and reconstruction results with different methods. From top to bottom: ground truth; results with Tikhonov regularization; results with I-FETV regularization and results with I-GTV regularization.}
\label{fig:f12}
\end{figure}

In the experiments with phantom data, only the central region is reconstructed in all the cases because the positions of sources and detectors lead to very low sensitivity away from the centre. It can be seen from the second row of Fig.~{\ref{fig:f12}}(c), that Tikhonov regularization over-smooths the reconstructed images which have much lower image contrast than the ground truth, especially in the first slice image. Artefacts are clearly observed near the source and detector areas. Even though total variation regularization can alleviate the over-smoothing effect caused by Tikhonov regularization, discretization methods still play an important role in the reconstruction performance. It should be noticed that I-FETV can alleviate the artefacts near to the sources and detectors but introduce some artefacts (staircase effect) in the non-anomaly area and does not preserve edges. I-GTV is seen to recover the anomaly with clear edges and high image contrast. It is interesting to compare these results to those of our previous work \cite{lu20181} where L$_1$ regularization was applied to the phantom data. Reconstructions by L$_1$ regularization were found to be over-sparsified and over-compact. In this work, TV regularization, which induces sparsity to the gradient of the solution, is seen to effectively alleviate the over-sparsifying effect of L$_1$ regularization and is therefore suitable for non-sparse coefficient distributions. We calculate the four evaluation metrics in the volume of illumination (Table~\ref{tb:compa1}) and these support the same conclusions. Similar localization errors are obtained by the different methods with only 1mm difference. Comparing to I-FETV, I-GTV can obtain the highest average contrast and PSNR values with similar relative recovered volume. 

\begin{table}[ht]
\centering
\caption{Evaluation of different methods for reconstruction on a tissue-simulating phantom.}
\label{tb:compa1}
\resizebox{\columnwidth}{!}{
\vspace{-10pt}
\begin{tabular}{|c|c|c|c|c|}
\hline
& Localization error / mm & Average contrast / - & PSNR / -   & Relative recovered volume /  \% \\ \hline
Tikhonov   & 2.90      & 0.74      & 13.74  & 40   \\ \hline
I-FETV    & 2.81      & 0.69      & 14.77  & 48   \\ \hline
I-GTV    & 3.16      & 0.79      & 16.71  & 46   \\ \hline
\end{tabular}}
\end{table}

\section{Conclusion}

In this paper, we introduce finite element and graph representations to discretize the TV regularization term in DOT reconstruction. Isotropic and anisotropic variants of the TV regularization are also investigated and compared between their FE- and graph-based implementations. The ADMM-based algorithms are proposed for each TV-regularized inverse problem. Experiments on the anisotropic TV regularization reveal that finite element representation yields the 'blocky' artefacts which is the designed in feature in the anisotropic TV regularization. However the graph representation favors the underlying shape of the region of interest so that the 'blocky' artefacts are not realized. Graph discretization on anisotropic TV regularization can adapt to the ground truth solution, but is not a good way to preserve anisotropy. 

Numerical experiments on isotropic TV regularization illustrate that, comparing to Tikhonov regularization, TV regularization can alleviate the over-smoothing effect of Tikhonov regularization and localize the anomaly by inducing sparsity of the gradient of the solution. These findings were tested on real experimental data using a tissue-simulating phantom. I-FETV does not perform well on low resolution reconstruction meshes because of the discrete nature of the finite element representation. Because the finite element representation operates on each element, the discretization becomes more accurate when the mesh resolution increases. I-GTV is shown to be more stable and robust to changes in mesh resolution because I-GTV is discretized on the graph directly, having no information of elements. Hence I-GTV can give more accurate discretization when the reconstruction mesh is a coarse mesh which is the usual case in DOT. However, I-FETV will outperform I-GTV when an reconstruction mesh with high resolution is available.

\section*{Acknowledgments}
This project has received funding from the European Union's Horizon 2020 Marie Sklodowska-Curie Innovative Training Networks (ITN-ETN) programme, under grant agreement no 675332, BitMap. We thank Daniel Lighter for providing  the experimental data used in Section 4.3.3. Supporting materials can be freely downloaded from \url{https://doi.org/10.25500/eData.bham.00000283}. 

\section*{Disclosures}
The authors declare that there are no conflicts of interest related to this article.


\end{document}